\documentclass{ieeeaccess}
\usepackage{cite}
\usepackage{here}
\usepackage{amsmath,amssymb,amsfonts}
\usepackage{algorithmic}
\usepackage{graphicx}
\usepackage{textcomp}
\usepackage{comment}
\usepackage{amssymb}
\usepackage{multirow}
\usepackage{autobreak}
\usepackage{amsmath,amssymb,amsfonts}
\usepackage{algorithmic}
\usepackage{graphicx}
\usepackage{textcomp}
\usepackage{graphics} % for pdf, bitmapped graphics files
\usepackage{epsfig} % for postscript graphics files
\usepackage{multirow}
\usepackage{colortbl}
\usepackage{amssymb}
\usepackage{amsmath}
\usepackage{color}
\usepackage{booktabs}
\usepackage{bm}
\usepackage{colortbl}
\usepackage{array}
\usepackage{autobreak}
\usepackage{mathtools}
\usepackage{threeparttable}

\def\BibTeX{{\rm B\kern-.05em{\sc i\kern-.025em b}\kern-.08em
    T\kern-.1667em\lower.7ex\hbox{E}\kern-.125emX}}
\begin{document}
\history{Date of publication xxxx 00, 0000, date of current version xxxx 00, 0000.}
\doi{10.1109/ACCESS.2017.DOI}
%methods~, model for imitation leanring
\title {An Independently Learnable Hierarchical Model for Bilateral Control-Based Imitation Learning Applications}%learning from unstructured demonstration
\author{\uppercase{Kazuki Hayashi}\authorrefmark{1},
\uppercase{Sho Sakaino\authorrefmark{2},\IEEEmembership{Member, IEEE}, and Toshiaki Tsuji}.\authorrefmark{3},
\IEEEmembership{Senior Member, IEEE}}
\address[1]{Faculty of Engineering, Information and Systems, University of Tsukuba, Ibaraki 305-8577, Japan (e-mail: s2020780@s.tsukuba.ac.jp)}
\address[2]{Department of Intelligent Interaction Technologies, University of Tsukuba, Ibaraki 305-8577, Japan (e-mail: sakaino@iit.tsukuba.ac.jp)}
\address[3]{Department of Electrical and Electronic Systems, Saitama University, Saitama 338-8570, Japan (e-mail: tsuji@ees.saitama-u.ac.jp)}
\tfootnote{
This work was partly supported by the Adaptable and Seamless Technology Transfer Program through Target-driven R\&D (A-STEP) from
the Japan Science and Technology Agency (JST) Grant Number JPMJTR20RG and the Japan Society for the Promotion of Science by a Grant-in-Aid for
Scientific Research (B) under Grant 21H01347.}

\markboth
{Author \headeretal: Preparation of Papers for IEEE TRANSACTIONS and JOURNALS}
{Author \headeretal: Preparation of Papers for IEEE TRANSACTIONS and JOURNALS}

\corresp{Corresponding author: Kazuki Hayashi (e-mail: s2020780@s.tsukuba.acjp).}

\begin{abstract}
Recently, motion generation by machine learning has been actively researched to automate various tasks. Imitation learning is one such method that learns motions from data collected in advance. 
However, executing long-term tasks remains challenging.
Therefore, a novel framework for imitation learning is proposed to solve this problem. 
The proposed framework comprises upper and lower layers, where the upper layer model, whose timescale is long, and lower layer model, whose timescale is short, can be independently trained. In this model, the upper layer learns long-term task planning, and the lower layer learns motion primitives.
The proposed method was experimentally compared to hierarchical RNN-based methods to validate its effectiveness.
Consequently, the proposed method showed a success rate equal to or greater than that of conventional methods.
In addition, the proposed method required less than 1/20 of the training time compared to conventional methods.
Moreover, it succeeded in executing unlearned tasks by reusing the trained lower layer.
\end{abstract}


\titlepgskip=-15pt

\maketitle

%背景(作業の自動化)、機械学習によるロボットの動作生成、模倣学習、自己回帰、階層化

\section{Introduction}
\label{sec:intro}
Recently, motion generation based on machine learning has been studied to automate various processes using robots.
For robotic automation, two methods, namely, reinforcement learning and imitation learning, have been mainly researched.

In the first method, the robots learn motions autonomously by repeating trials. 
However, this method needs many trials to acquire the motion policies~\cite{levine}. 
To solve this problem, Sim2real, in which robots repeat trials in a simulation environment instead of the real world, was proposed \cite{sim2real}. 
Although this method is effective, robots sometimes cannot behave properly in practice because of the gap between the simulated and real environments \cite{sim2real_prob}.

On the other hand, in imitation learning, robots acquire motion from expert data collected in advance\cite{survey,survey2}. Therefore, this requires significantly fewer demonstrations than that of reinforcement learning.
Subsequently, imitation learning is practical for generating motions in robots. Moreover, imitation learning with force information enables robots to perform various tasks with contacts\cite{f0,f1,f2,f3,f4}.
In particular, neural networks (NNs) are promising techniques for inferring complicated motions. Recurrent NNs (RNNs), which enable time-series inference, are effective for generating robotic motions\cite{rnn-im}.
However, imitation learning still has problems when executing long-term tasks \cite{survey0,long-term,relay}. 

Long short-term memory (LSTM) has been proposed to improve the long-term inference of RNNs but its performance remains insufficient.
In addition, when multiple sensors are used simultaneously, the different sampling periods of the sensors can be a problem.
To address this problem, two methods for dividing a large task into multiple subtasks have been studied.

The first approach treats many subtasks as a monolithic task. This method does not require subtask segmentation.
Although RNN has limitations with respect to the length of the time series, it can be solved by introducing a hierarchical structure\cite{hrnn,hrnn2}.
Saito \textit{et al.} succeeded in executing tasks, such as wiping on a complex surface with a multi-timescale RNN (MTRNN)\cite{ogata1}.
MTRNN is a type of hierarchical RNN that consists of three nodes that update every time constant.
However, to consider longer tasks and operations in various environments, an enormous number of demonstrations are needed\cite{hil1}.

The second approach is to divide a long-term task into several subtasks or primitives. After segmentation, it is possible to execute various tasks by rearranging or re-utilizing them.
This method needs fewer demonstrations compared to the first approach.
However, the segmentation of the task sometimes needs human intervention\cite{hil2}.
Although many methods for automating this process have been researched \cite{segment1,segment2,segment3,segment4,segment5,segment6}, most of them require strong mathematical assumptions or prior knowledge about the task\cite{prob_segment}.

Considering that the behavior of a mechanical system is governed by position and force, we can infer that the segmentation of tasks will be easier if the position and force information can be accurately reproduced. In fact, past research has shown that extracting movement primitives by utilizing the interaction force results in better generalization abilities\cite{prim1}.
Subsequently, we proposed bilateral control-based imitation learning that enables robots to execute tasks requiring force adjustment and fast behavior\cite{bilate-im1,bilate-im2,bilate-im3,sm2sm,s2sm}.
Bilateral control is a remote-control technique for leader and follower robots with force feedback \cite{sm2sm}.
The force information collected with bilateral control is helpful for learning the movement primitive.
Therefore, we propose a hierarchical imitation learning method for bilateral control-based imitation learning, which has the merits of both abovementioned approaches. 
In other words, our method does not require explicit task segmentation, instead few demonstrations are required.

In our proposed method, two types of LSTM, an upper layer LSTM and a lower layer LSTM, are used.
In the proposed model, the inference of the slow-moving higher layer can be regarded as almost unchanged from perspective of the fast-moving lower layer.
In the proposed method, the upper layer LSTM is trained to learn long-term task planning from the time series of the follower state.
Then, the lower layer LSTM is trained to learn motion primitives from the states of the leader and follower robots.
In the training stage of the lower layer LSTM, the force information brought by bilateral control improves the generalization ability and adaptability to environmental changes.
Note that conventional hierarchical imitation learning cannot accurately reproduce position and force responses in the frequency domain; thus, it is necessary to maintain a mechanism where the upper and lower layers feedback each other to correct errors. However, in our bilateral control-based imitation learning framework because both position and force responses can be accurately reproduced in the frequency domain, the feedback from the lower layer to the upper layer is unnecessary. Therefore, the upper and lower layers can be trained independently.
The features of our proposed method are summarized as follows.

\begin{enumerate}
   \item The upper layer can learn long-term task planning without task segmentation.
   \item The lower layer LSTM learns motion primitives by utilizing the force information collected by bilateral control.
   \item The upper layer LSTM and lower layer LSTM are independently trained. This feature reduces training time. Additionally, the lower layer infers motion primitives and allows robots to perform various tasks.
   \item The proposed method retains characteristics of bilateral control-based imitation learning, such as the fast movement, force adjustment, and adaptivity to environmental changes. 
   \item It enables robots to execute longer-term tasks.
\end{enumerate}

In this study, the proposed method was compared to three conventional hierarchical RNN types: MTRNN\cite{mtrnn}, Clockwork RNN\cite{cwrnn}, and Fast-Slow RNN\cite{fsrnn}. 
%In the experiments, four tasks were conducted. 
In the experiments, our proposed method showed a performance equal to or better than that of the conventional methods.
Moreover, training with our method required less than 1/20 of the training time compared to conventional methods.
In addition, the lower LSTM can be reused for an unlearned task by writing unlearned characters.
%our method succeeded in writing unlearned characters by re-utilizing the already trained lower layer LSTM.

The remainder of this paper is structured as follows. 
Section II introduces the procedure of bilateral control-based imitation learning. This section consists of data collection (demonstration) with bilateral control, preprocessing, and training with the collected data.
In section III, our proposed method and conventional hierarchical RNNs are described. 
In section IV, the experimental procedures and results are explained. Here, the types of conventional hierarchical RNNs are compared to our proposed method.
In section V, the conclusion of this study and future research directions are described.

\section{Related work}%bilate based imitation 
In imitation learning, robots acquire skills from demonstrations. 
On the other hand, bilateral control is the teleoperation method between two robots, the leader manipulated by an expert and the follower who operates in the workspace. During the bilateral control, angles and torques of the follower and the leader are synchronized. Moreover, collecting demonstrations with bilateral control has several advantages. 

Firstly, it enables robots to operate fast. As described above, the state of the follower and that of the leader are synchronized.  In other words, the command value for the follower is the state of the leader.
Thus, robots can operate quickly by using the predicted leader's state as the command value. On the other hand, this is impossible when bilateral control is not used because the appropriate command value is unavailable. In addition, this trait helps variable speed motion generation\cite{variable}.

Second, it allows the robot to operate with the right adjustment of force. In bilateral control, the torque responses of the follower and the leader are synchronized and the law of action and reaction is established between them. Here, the leader measures the action force, and the follower observes the reaction force. Thus, collecting demonstrations with bilateral control enables robots to imitate the force adjustments\cite{bilate-im3}.

However, in the past, our model tended to be unstable in autonomous operation because it was trained with teacher forcing\cite{sm2sm}. To solve this problem, Sasagawa et al. proposed the FL2FL model to train models with scheduled sampling. 
Although only the follower performs tasks in autonomous operation,  the FL2FL needs two inputs, the states of the follower and the leader. 
Thus, the predicted leader state and the current follower's state are used as the inputs in using the FL2FL model\cite{sm2sm}.
However, this procedure causes the covariate shift and destabilizes the autonomous operation. Hence the F2FL model was proposed to solve it\cite{s2sm}.

\section{Bilateral Control-Based Imitation Learning}
\label{sec:system}
In general imitation learning approaches, such as direct teaching, only one robot's responses are available and next step responses are treated as commands.
However, because the commands were substituted for the responses, only low-frequency operations could be realized if responses and commands could be assumed to be consistent.
Contrarily, commands for a follower are responses of a leader during bilateral control.
Therefore, both commands and responses for a follower are available in bilateral control-based imitation learning\cite{sm2sm}.
Therefore, our bilateral control-based imitation learning enables robots to operate quickly owing to using the predicted leader state as a command value.
Moreover, the force information is obtained as the torque response in our method. Therefore, our method is suitable for executing many tasks that require force information and adaptivity to environmental changes.

\subsection{Data Collection With Bilateral control}
4ch bilateral control was used to collect training datasets\cite{bilate-im3}. 
In this study, experts manipulate the leader, while the follower is teleoperated by the leader.
This is obtained by the synchronizing the positions and feedback of the forces between the leader and follower robots. 
The control goals of the bilateral control are summarized as follows:

\begin{eqnarray}
\theta_l^{res} - \theta_f^{res} &=& 0 \label{4ch-goal1} \\
\tau_l^{res} + \tau_f^{res} &=& 0 . \label{4ch-goal2}
\end{eqnarray}
Here, the superscripts $res$ represent the response value, and $l$ and $f$ indicate the leader and follower, respectively.
Moreover, $\theta$ and $\tau$ denote the angle and torque, respectively.
During data collection, the control period of the robots was 1 ms.

The details can be found in our previous research\cite{s2sm,sm2sm}. Note that the leader is referenced as the master, and the follower is called the slave in these studies.

\subsection{Preprocessing}%decimation and maxmin_scaling
After collecting the motion data, the sampling rate of the training data was decimated to 20 ms, and the rejected data were reused for data augmentation\cite{augment}.
Afterward, min-max normalization was executed on the decimated data.
Here, the states of the follower and leader in each batch at the $t$-th time step are defined as follows:
\begin{eqnarray}
L_t  &=& [\theta_l(t),\dot\theta_l(t),\tau_l (t)]  \\
F_t  &=& [\theta_f(t),\dot\theta_f(t),\tau_f (t)].
\end{eqnarray}

\subsection{F2L Model}%Slave2Master
\label{f2l}
This is the basic model for bilateral control-based imitation learning, which is the same as the S2M model described in \cite{sm2sm}.

\subsubsection{Training}
The model is trained to predict the next leader's response value $L_{t+1}$ from the current follower's response value $F_t$, as depicted in Fig.~\ref{fig:s2m-train}.
\subsubsection{Autonomous operation}
As presented in Fig.~\ref{fig:s2m-auto}, the trained model receives the current follower's state and predicts the next leader's state, which is used as the command value for the follower during its autonomous operation.
Although the F2L model succeeded in some tasks, it is not suitable for executing long-term tasks. This is because the F2L model does not consider the accumulation of prediction errors during autonomous operations.

\subsection{FL2FL Model (For autoregressive learning)}%slave master 2 slave master
\label{fl2fl}
This model is the same as the SM2SM model described in \cite{sm2sm}.
To solve the problem of the F2L model, the FL2FL model was proposed.

\subsubsection{Training}
\label{auto}
This model is trained to predict the next states of the follower and leader $[F_{t+1},L_{t+1}]$ from the current states of the follower and leader $[F_{t},L_{t}]$.
In addition, the predicted states of the follower and leader $[\hat{F}_{t+1},\hat{L}_{t+1}]$ are used as the model input in the next step, as depicted in Fig.~\ref{fig:sm2sm-train}.
This is called autoregressive learning or scheduled sampling, and Sasagawa \textit{et al.} showed that it is effective for generating motions for long-term tasks\cite{sm2sm}.

\subsubsection{Autonomous operation}
During autonomous operations, the leader's response value does not exist because only the follower robot operates alone.
Therefore, the predicted state of the leader in the previous step and the current state of the follower were used as the model input, as depicted in Fig.~\ref{fig:sm2sm-auto}.
Similar to the F2L model, the predicted value of the leader's response was treated as the command for the follower.
However, using the predicted leader state as the model input sometimes causes a covariate shift\cite{covariate}.

\subsection{F2FL Model (For autoregressive learning)}%Slave 2 slave and Leader
\label{f2fl}
This model is the same as the S2SM model described in \cite{s2sm} and is regarded as an improved type of the FL2FL model.
To solve the covariate shift problem of the FL2FL model, the F2FL model proposes eliminating the leader's responses from the inputs.
\subsubsection{Training}
The F2FL model is trained to predict the next states of the follower and leader $[F_{t+1},L_{t+1}]$ from the current state of the follower $F_t$.
Additionally, the predicted state of the follower $\hat{F}_{t+1}$ is used as the next input of the model, as presented in Fig.~\ref{fig:s2sm-train}.
\subsubsection{Autonomous operation}
This model receives the current state of the follower and predicts the next states of the follower and leader.
Moreover, the predicted response of the leader is used as the command value for the follower, as depicted in Fig.~\ref{fig:s2sm-auto}.
Note that the responses of the leader are not required to reduce the covariate shit effect.
% 図の挿入

\Figure[t!](topskip=0pt, botskip=0pt, midskip=0pt)[width=8cm]{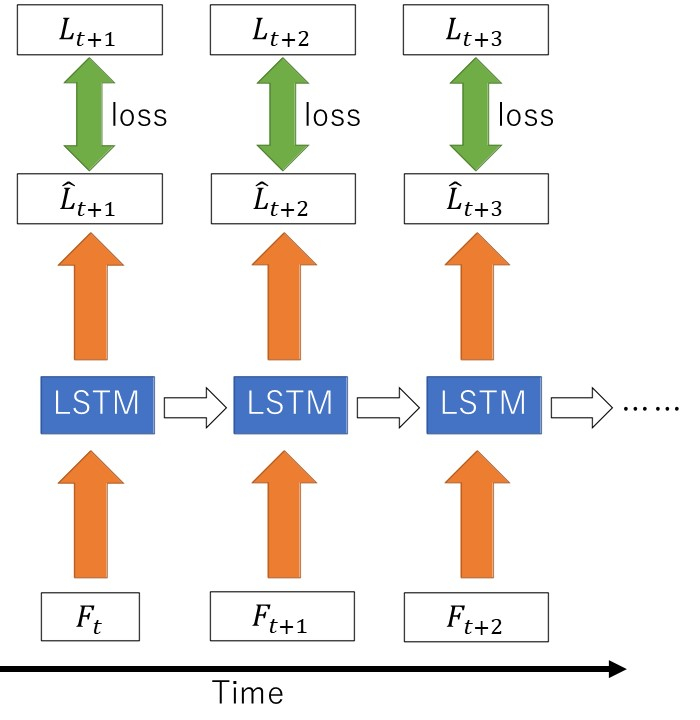}
{Training of the F2L model. \label{fig:s2m-train}}
\Figure[t!](topskip=0pt, botskip=0pt, midskip=0pt)[width=8cm]{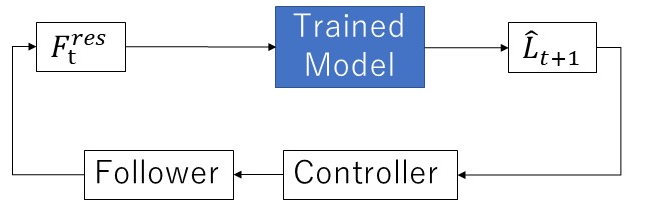}
{Autonomous operation using the F2L model. \label{fig:s2m-auto}}

\Figure[t!](topskip=0pt, botskip=0pt, midskip=0pt)[width=8cm]{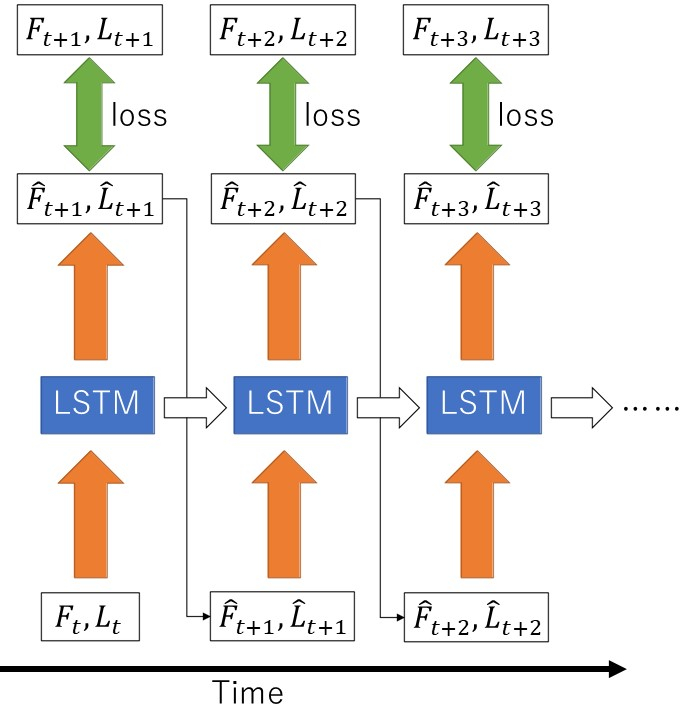}
{Training of the FL2FL model. Note that the predicted states of the follower and leader $\hat{L}_t,\hat{F}_t$ are used as the input in the next step. \label{fig:sm2sm-train}}
\Figure[t!](topskip=0pt, botskip=0pt, midskip=0pt)[width=8cm]{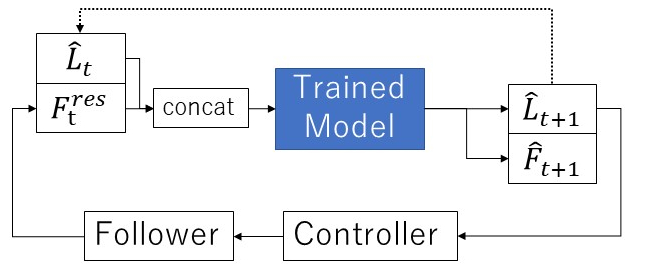}
{Autonomous operation using the FL2FL model. Note that the leader state predicted previously is used as the next input because only the follower operates.\label{fig:sm2sm-auto}}

\Figure[t!](topskip=0pt, botskip=0pt, midskip=0pt)[width=8cm]{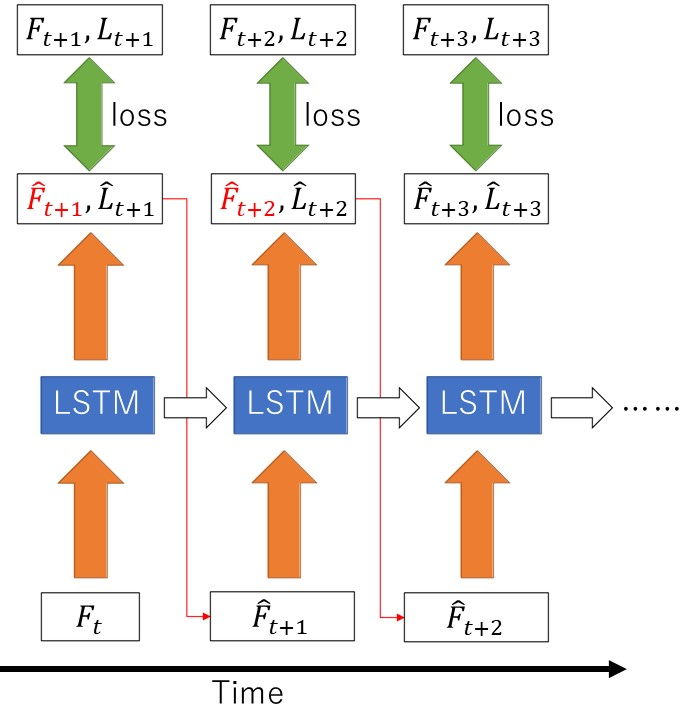}
{Training of the F2FL model. Note that only the predicted follower state is used as the input in the next step.
\label{fig:s2sm-train}}
\Figure[t!](topskip=0pt, botskip=0pt, midskip=0pt)[width=8cm]{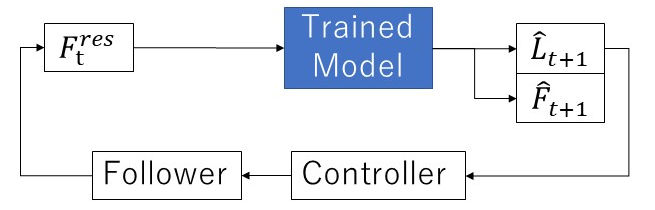}
{Autonomous operation using the F2FL model. \label{fig:s2sm-auto}}

\subsection{Detailed explanation of autoregressive learning}%Slave 2 slave and Leader
\subsubsection{Introduction of scheduled sampling}
As explained in section \ref{f2l}, the model input is the follower state at time step t $(F_t)$, and its output is the next leader state $(L_{t+1})$ used in training the F2L model. Note that the model input is always the follower state of the training data. This training method is called "teacher forcing" because the model input is always the values of the teacher (training) data.
Ranzato \textit{et al.} asserted that training with teacher forcing decreases robustness because the model never trains on its own error\cite{autoreg1}. 
To solve this problem, scheduled sampling was proposed\cite{autoreg2}. 
In training models with scheduled sampling, the model output at the previous step is used as the input, as explained in sections \ref{fl2fl} and \ref{f2fl}.

\subsubsection{Qualify the autoregressive learning's effectiveness}
\label{qualify}
In summary, the model trained with scheduled sampling is expected to generate appropriate behavior even if the input is slightly different from the probabilistic distribution of the training data.
Hence, we studied whether the generated motion was similar to the training data by comparing their probability distributions.

The Pearson (PE) divergence, which measures the difference between two probabilistic distributions $p(Y), p'(Y)$, is defined as follows:

\begin{equation}
PE(P||P') :=\frac{1}{2} \int p'(Y) \left( \frac{p(Y)}{p'(Y)}-1 \right)^2 dY .
\end{equation}
According to \cite{ulsif1}, the similarity is defined as follows:
\begin{equation}
PE(P_t||P_{t+n}) + PE(P_{t+n}||P_t) .
\end{equation}
Here, $P_t$ is the distribution of samples in the Hankel matrix $\mathit{Y_t}$, which was obtained from the time-series $[y_1, ...,y_T]$. This index refers to the difference in the distribution of a single time series at different time steps and is used for change-point detection.
Therefore, let us define $P'_t$ as the distribution of samples in the Hankel matrix $\mathit{X_t}$, which was obtained from the time-series $[x_1, ...,x_T]$.

Moreover, the similarity of the distributions of two time-series data $[x_1, ...,x_T],[y_1, ...,y_T]$ in the same time step $t$ is defined as follows:
\begin{equation}
PE(P_t||P'_{t}) + PE(P'_{t}||P_t). \label{qualify2}
\end{equation}
We used this index to qualify the effectiveness of autoregressive learning.

%バイラテベース模倣学習や自己回帰について

\section{Types of neural networks for hierarchical imitation learning}
\label{sec:system}

\subsection{Basic LSTM structure}\label{sec:lstm}
As depicted in Fig.~\ref{fig:lstm}, our basic LSTM consists of an input layer (LSTM cell), a hidden layer (LSTM cell), and an output layer (fully connected layer).
Moreover, each LSTM cell had the same node size. Therefore, all hyperparameters were the node size and the number of layers.
\Figure[t!](topskip=0pt, botskip=0pt, midskip=0pt)[width=8.5cm]{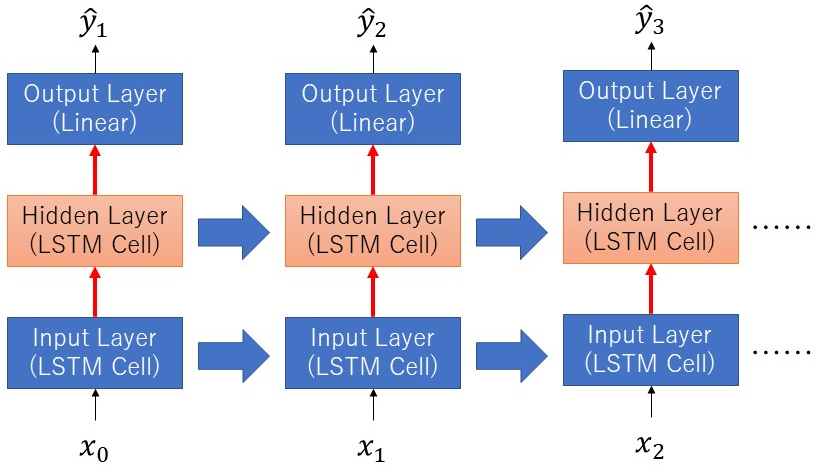}
{Diagram of our basic LSTM when the number of hidden layers is one. It consists of multiple LSTM cells with the same node size and a fully connected layer.\label{fig:lstm}}

\subsection{Proposed method I (Ordinary model)}
In the proposed method I, the upper and lower layers are trained independently.
\subsubsection{Training of the upper layer LSTM}
The upper LSTM is trained to predict the state of the follower 20 steps later ${F}_{t+20}$ from the current follower's state $F_t$.
In training, note that the predicted follower state $\hat{F}_{t+20}$ is used as the input in the next step, as presented in Fig.~\ref{fig:prop1-upper} (autoregressive learning).
As well as in section \ref{sec:lstm}, the LSTM model with the configuration shown in Fig.~\ref{fig:lstm} was used.

\subsubsection{Training of the lower layer LSTM}
The lower layer LSTM is trained to predict the next states of the follower and leader $[\hat{F}_{t+1},\hat{L}_{t+1}]$ from the current state of the follower $F_t$ (and the leader state $L_t$) and the future follower's state $F_{t+20}$, as shown in Fig. \ref{fig:prop1-lower}. Note that the future follower state $F_{t+20}$ is updated every 20 steps. In addition to the training of the FL2FL, the predicted states $[\hat{F}_{t+1},\hat{L}_{t+1}]$ are used as the input in the next step.
In summary, the lower layer is the F2FL or FL2FL model that uses $F_{t+20}$ as its input.
As well as in section \ref{sec:lstm}, the LSTM model with the configuration shown in Fig.~\ref{fig:lstm} was used.

\subsubsection{Autonomous operation}
Fig.~\ref{fig:prop1-auto} demonstrates the procedure for autonomous operation with the follower.
In autonomous operation, the upper layer receives the follower response $F_t^{res}$ and outputs its future goal state $\hat{F}_{t+20}$.
Then, the lower layer predicts the next states of the follower and leader $[\hat{F}_{t+1},\hat{L}_{t+1}]$ from the follower response $F_t$ and its future goal state $\hat{F}_{t+20}$. 
Similar to other models, the predicted leader response at the next time step $\hat{L}_{t+1}$ is used as the command value. Moreover, the future goal state $\hat{F}_{t+20}$ is updated every 20 time steps.

\subsection{Proposed method II (Angle model)}
\subsubsection{Training of the upper layer LSTM}
The upper layer LSTM is trained to predict the future angles of the four followers $\lbrack \theta_f (t+10),\theta_f (t+20),\theta_f (t+30),\theta_f (t+40) \rbrack$ from the current follower angle $\theta_f (t)$, as shown in Fig.~ \ref{fig:prop2-upper}. Furthermore, the predicted angle $\hat{\theta}_f (t+10)$ is used as the next input (autoregressive learning).
As well as in section \ref{sec:lstm}, the LSTM model with the configuration shown in Fig.~\ref{fig:lstm} was used.

\subsubsection{Training of the lower layer LSTM}
The lower layer LSTM is trained to predict the next states of the follower and leader $[\hat{F}_{t+1},\hat{L}_{t+1}]$ from the current state of the follower $F_t$ (and the leader state $L_t$) and the future follower angles $\lbrack \theta_f (t+10),\theta_f (t+20),\theta_f (t+30), \theta_f (t+40) \rbrack$, as depicted in Fig. \ref{fig:prop2-lower}.
Note that the future follower angles are updated every 10 steps. In addition to the training of the FL2FL, the predicted states $[\hat{F}_{t+1},\hat{L}_{t+1}]$ are used as the input in the next step.
In summary, the lower layer is the F2FL or FL2FL model that uses the future goal angles $\lbrack \theta_f (t+10),\theta_f (t+20),\theta_f (t+30), \theta_f (t+40) \rbrack$ as its input.
As well as in section \ref{sec:lstm}, the LSTM model with the configuration shown in Fig.~\ref{fig:lstm} was used.

\subsubsection{Autonomous operation}
Fig. \ref{fig:prop2-auto} demonstrates the procedure for autonomous operation with the follower.
In autonomous operation, the upper layer receives the follower's angular response $\theta_f^{res}(t)$ and outputs its future goal angles $\lbrack \hat{\theta}_f (t+10), \hat{\theta}_f (t+20),\hat{\theta}_f (t+30), \hat{\theta}_f (t+40) \rbrack$.

Then, the lower layer predicts the next states of the follower and leader $[\hat{F}_{t+1},\hat{L}_{t+1}]$ from the follower's response $F_t^{res}$ and its future goal angles.
Similar to other models, the predicted leader response at the next time step $\hat{L}_{t+1}$ is used as the command value. 
Moreover, the future goal angles $\lbrack \hat{\theta}_f (t+10), \hat{\theta}_f (t+20),\hat{\theta}_f (t+30), \hat{\theta}_f (t+40) \rbrack$ are updated every 10 time steps.

\Figure[t!](topskip=0pt, botskip=0pt, midskip=0pt)[width=8.5cm]{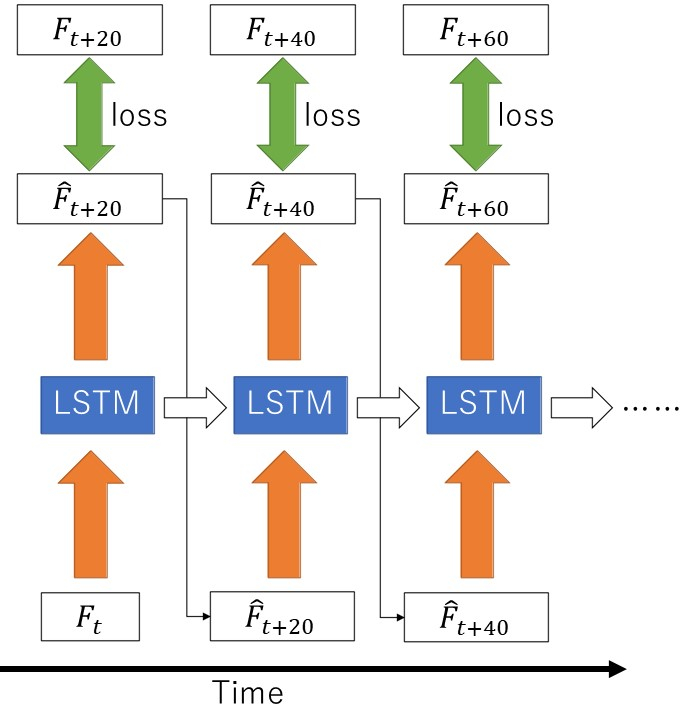}
{Training of the upper layer of the proposed method I. The timescale of the upper layer is 20 steps in the proposed method I.\label{fig:prop1-upper}}
\Figure[t!](topskip=0pt, botskip=0pt, midskip=0pt)[width=8.5cm]{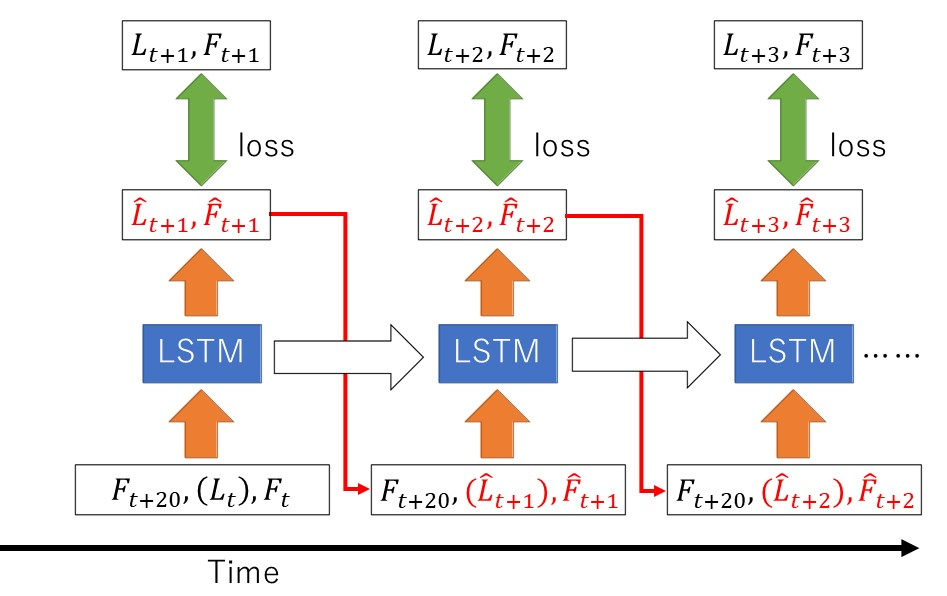}
{Training of the lower layer of the proposed method I. Note that the future follower state $F_{20}$ is updated every 20 time steps.\label{fig:prop1-lower}}
\Figure[t!](topskip=0pt, botskip=0pt, midskip=0pt)[width=8.5cm]{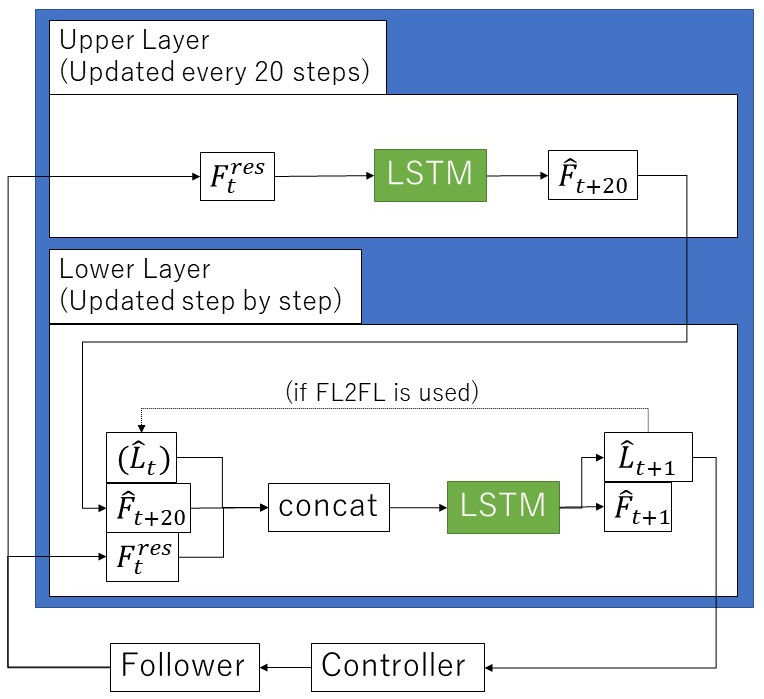}
{Autonomous operation with the proposed method I. Note that the future follower state $\hat{F}_{t+20}$ is updated every 20 time steps.\label{fig:prop1-auto}}

\Figure[t!](topskip=0pt, botskip=0pt, midskip=0pt)[width=10cm]{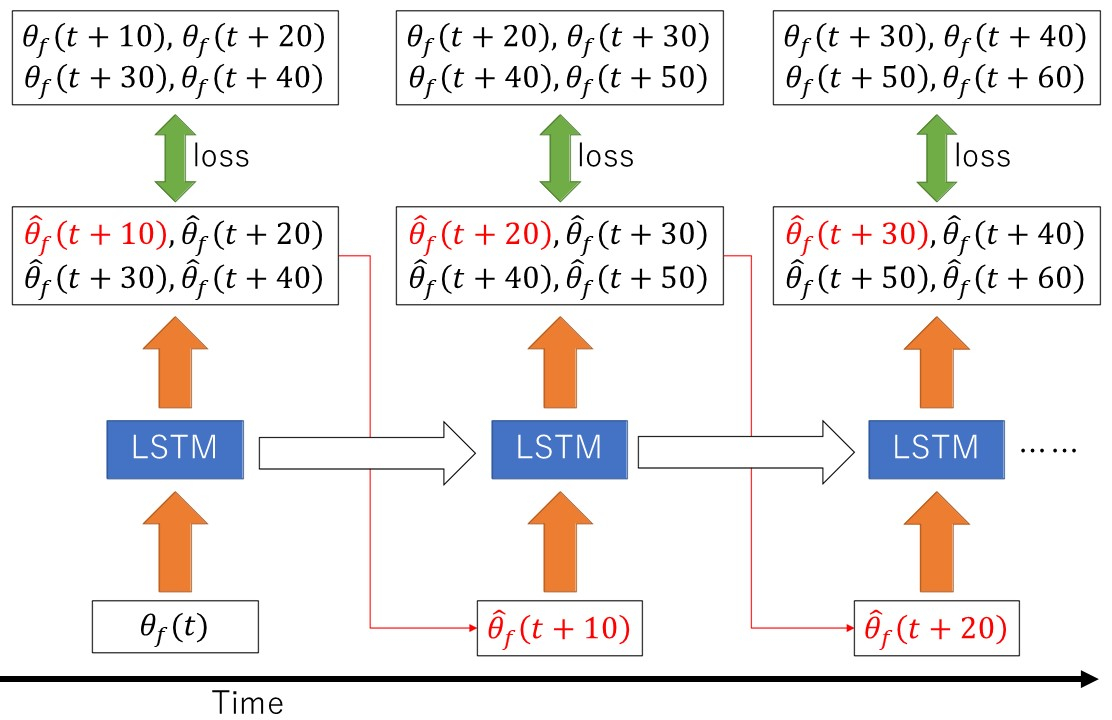}
{Training of the upper layer of the proposed method II. The timescale of the upper layer is 10 steps in the proposed method II.\label{fig:prop2-upper}}
\Figure[t!](topskip=0pt, botskip=0pt, midskip=0pt)[width=10cm]{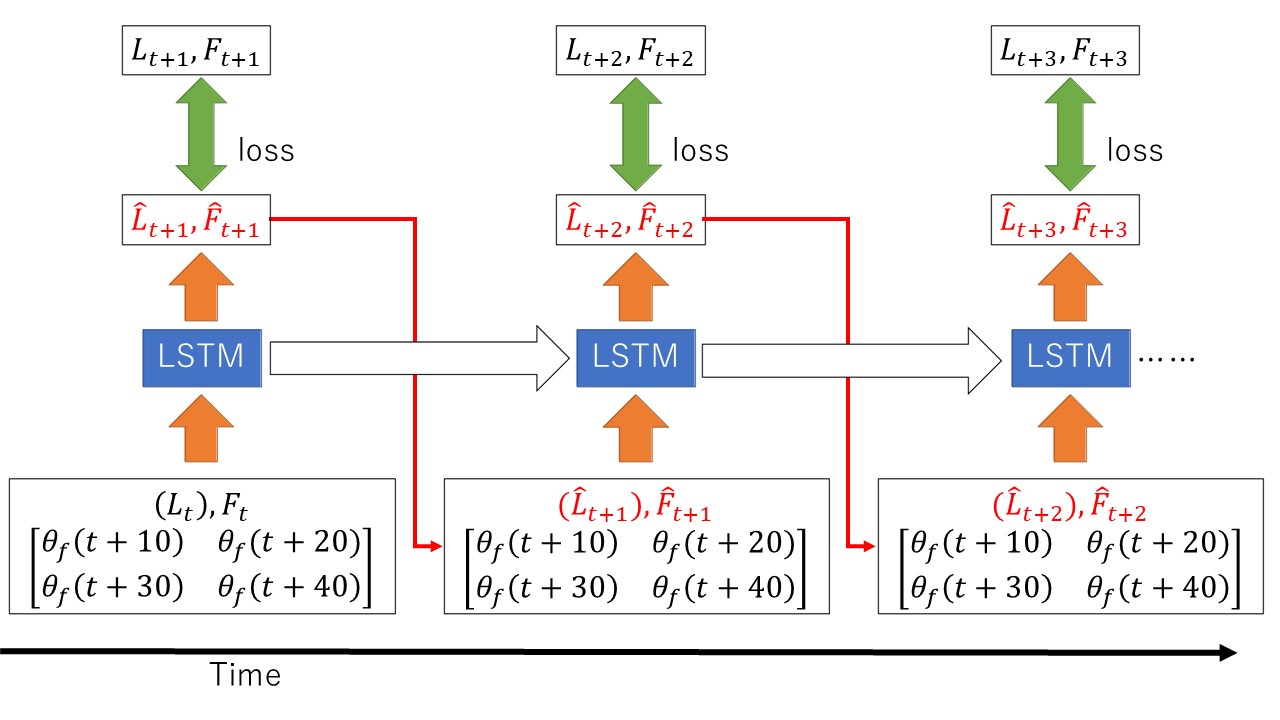}
{Training of the lower layer of the proposed method II. Note that the future follower state $\theta_f (t+10),\theta_f (t+20),\theta_f (t+30),\theta_f (t+40)$ is updated every 10 time steps.\label{fig:prop2-lower}}
\Figure[t!](topskip=0pt, botskip=0pt, midskip=0pt)[width=10cm]{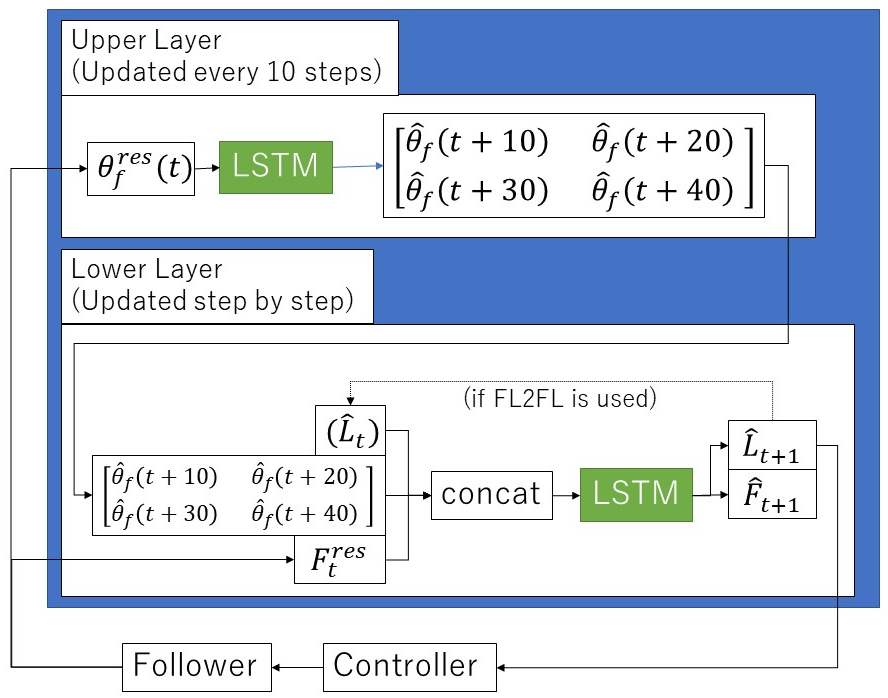}
{Autonomous operation using proposed method II. Note that the future follower state $[\hat{\theta}_f(t+10),\hat{\theta}_f(t+20),\hat{\theta}_f(t+30),\hat{\theta}_f(t+40)]$ is updated every 10 time steps.\label{fig:prop2-auto}}

%以降、比較手法
\subsection{Comparison methods}
\subsubsection{Fast-Slow RNN (FS-RNN)}
Fast-Slow RNN is a hierarchical RNN type \cite{fsrnn}. As presented in Fig.~\ref{fig:fsrnn}, it consists of two types of RNN cells, namely, fast and slow cells.
During the forward calculation of FS-RNN, the fast RNN cell sends its hidden state to the slow RNN cell once every few steps. 
When the slow RNN cell receives the hidden state, the slow RNN cell returns the hidden state to the fast RNN cell.
Because of this procedure, the fast RNN cell learns the short-term components, while the slow RNN cell learns the long-term dependencies.
In our experiments, two fully connected (FC) layers were attached to the slow cell, as shown in Fig.~\ref{fig:fsrnn}.
This is because short-term information that fast cell stores are required to control robots.
The node size was the only hyperparameter because the node sizes of layers should be equal. Moreover, the state of the upper layer was updated every 20 steps as well as the proposed method I.

\subsubsection{Clockwork RNN (CW-RNN)}
Clockwork RNN is  a hierarchical RNN type as well \cite{cwrnn}.
As depicted in Fig.~\ref{fig:cwrnn}, the hidden units are divided into modules in the Clockwork RNN.
Moreover, the $i$-th module is updated every $2^i$-th time-step. 
Owing to this structure, Clockwork RNN is computationally efficient and can learn long-term dependencies.
Moreover, hyperparameters were the node size and the number of divided modules.

\subsubsection{Multiple Timescale Recurrent Neural Network (MTRNN)}
As presented in Fig. \ref{fig:mtrnn}, MTRNN is a type of hierarchical RNN that is structured by three types of nodes with different time constants \cite{ogata1,mtrnn}.
These are slow context (Cs) nodes, fast context (Cf) nodes, and input/output (I/O) nodes.
The Cs nodes have a longer time constant and are expected to learn the long-term dependencies.
On the other hand, Cf nodes can learn motion primitives because they have a shorter time constant.
Owing to these three types of nodes, MTRNN enables robots to perform long-term tasks.

In the forward calculation, the state of the $i$-th neuron at time step $t$ is calculated as follows:
\begin{equation}%internal state of the neuron i at time step t is calculated as 
\begin{split}
u_i(t) = \sigma \left \lbrace \left( 1- \frac{1}{\tau_i} \right) u_i (t-1) + \frac{1}{\tau_i} \left( \sum_{j \in N} w_{ij} x_j (t) \right) \right\rbrace .
\end{split}
\end{equation}
Here, $N$ denotes the number of neurons connected to neuron $i$. 
Moreover, $w_{ij}$ is the weight from neuron $j$ to neuron $i$, and $\sigma$ denotes the sigmoid activation function.
Then, the output value is calculated with the sigmoid function as follows:
\begin{equation}
y_i (t) = \sigma (u_{io} (t)), \label{mtrnn-eq}
\end{equation}
where, $u_{io}$ is the neuron at I/O nodes.
As shown in Fig.\ref{fig:mtrnn}, $i$-th neuron and N neurons connected to it are described in Table \ref{mtrnn-tab}.

\begin{table}[h]
\caption{Neurons that are connected to the $i$-th neuron.}
\label{mtrnn-tab}
\begin{tabular}{|l|l|}
\hline
neuron $i$    & $N$(neurons connected to neuron $i$)  \\ \hline
I/O   & Fast         \\ \hline
Fast & I/O, Fast, Slow \\ \hline
Slow & Fast, Slow    \\ \hline
\end{tabular}
\end{table}

Moreover, hyperparameters were the node sizes and time constants of the Cf and Cs nodes.

%\Figure[t!](topskip=0pt, botskip=0pt, midskip=0pt)[width=8cm]{fsrnn.jpg}
%{Architecture of a Fast-Slow RNN\cite{fsrnn}. Black arrows indicate the transfer of the hidden states.\label{fig:fsrnn}}
\Figure[t!](topskip=0pt, botskip=0pt, midskip=0pt)[width=8cm]{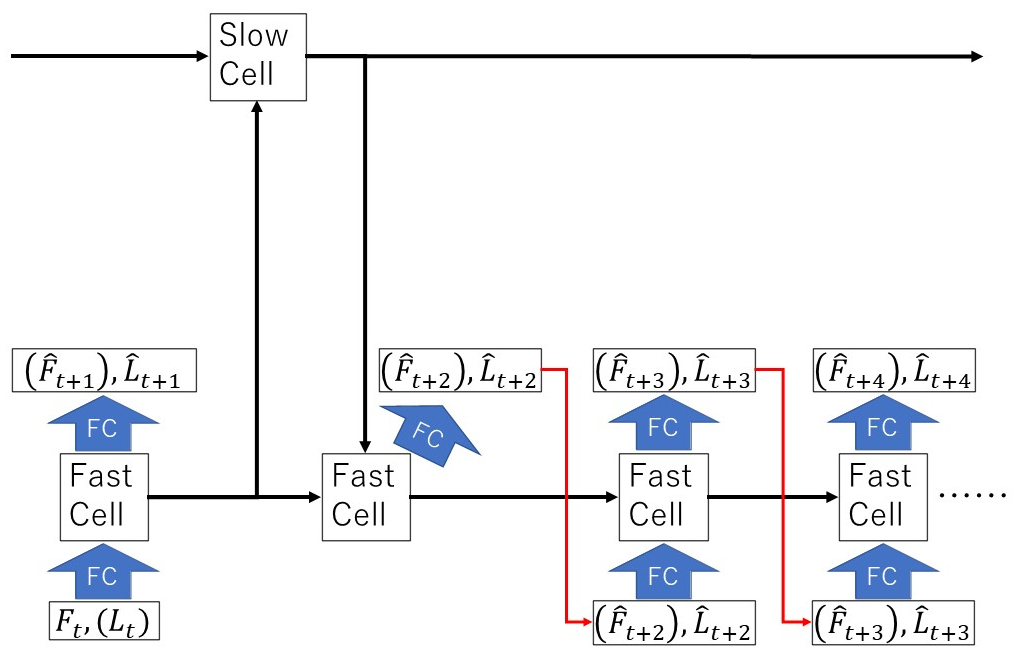}
{Training of the FL2FL or F2FL model with a fast-slow RNN. FC stands for a fully connected layer, and red arrows indicate the data flow of autoregressive learning adopted in the FL2FL or F2FL models.\label{fig:fsrnn}}

\Figure[t!](topskip=0pt, botskip=0pt, midskip=0pt)[width=8.5cm]{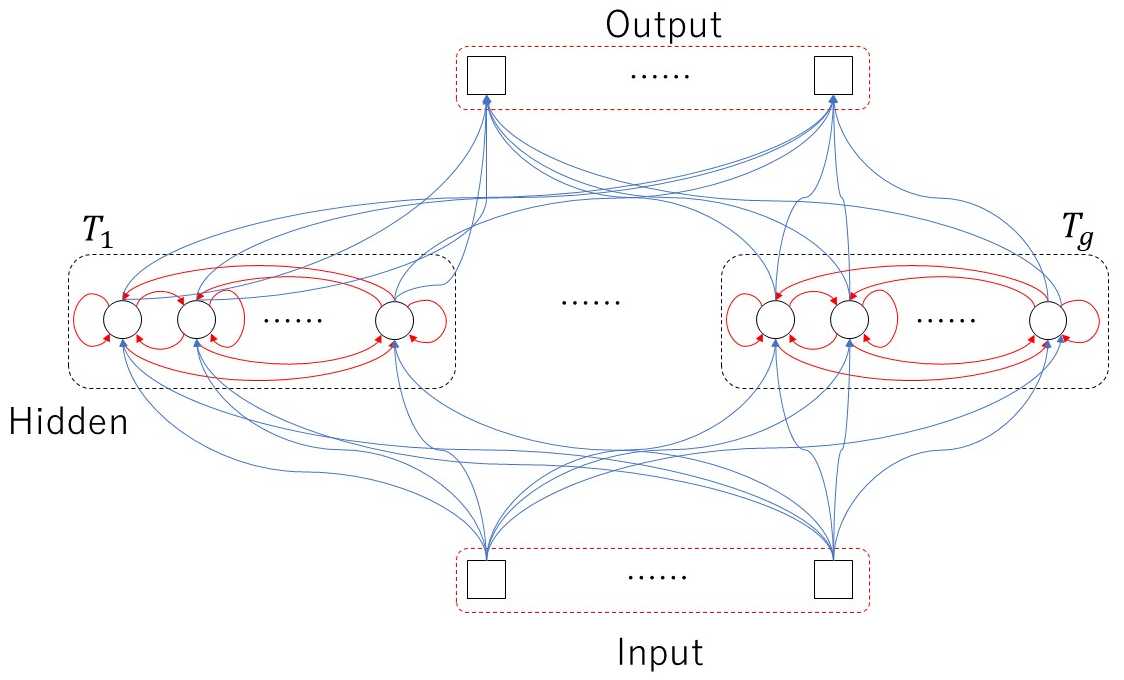}
{Diagram of a Clockwork RNN\cite{cwrnn}. The hidden unit is divided into $g$ modules and each module is updated in different cycles.\label{fig:cwrnn}}

\Figure[t!](topskip=0pt, botskip=0pt, midskip=0pt)[width=6cm]{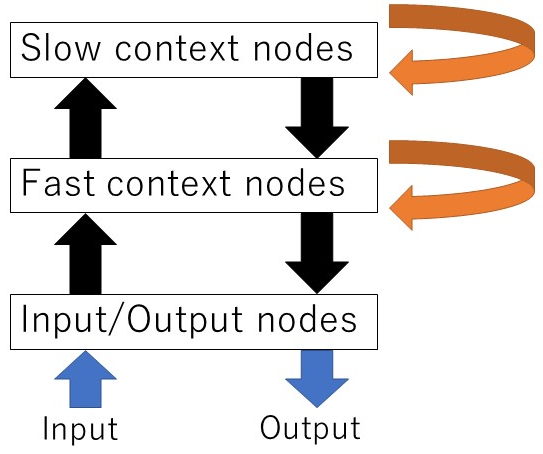}
{Diagram of a MTRNN. Each arrow indicates data flow among the nodes.\label{fig:mtrnn}}

%階層化について

\section{Experiment}
\label{sec:experiment}
In the experiment, four types of tasks were executed to demonstrate the effectiveness of the proposed method.
The proposed method was compared to CW-RNN, FS-RNN, MTRNN, and LSTM. 
Here, LSTM denotes our conventional method, which has no hierarchical structure \cite{bilate-im3,sm2sm,s2sm}.
During the experiments, the FL2FL and F2FL models were used inside the comparison methods, CW-RNN, FS-RNN, MTRNN, and LSTM.
Moreover, the F2L and FL2FL models were used in the MTRNN because the model input and output sizes should be equal when using the MTRNN.
In this study, three autoregression numbers 1, 5, 20 were used. 
These numbers indicate the frequency of scheduled sampling. For example, Fig. \ref{fig:s2sm-train} demonstrates the training case with an autoregression number of 3. 
In summary, we consider three steps of time-series data as one block of in the training. In other words, the value obtained from the training data is used once every three times.

Table \ref{pc-spec} lists the specifications of the computer that was used for the training and autonomous operation.

\begin{table}[]
\caption{The specifications of the computer that was used for the training and autonomous operation}
\begin{tabular}{|l|l|}
\hline
OS  & Ubuntu 18.04.6 LTS (64bit)              \\ \hline
RAM & 32 GB                                \\ \hline
CPU & AMD Ryzen 7 3700x 8-core procesor x 16  \\ \hline
GPU & NVIDIA GeForce RTX 2080 SUPER/PCIe/SSE2 \\ \hline
\end{tabular}
\label{pc-spec}
\end{table}

\subsection{Experiment 1 (Writing the character B)}
\subsubsection{Task Design}
In this experiment, a ballpoint pen was fixed to the follower, as depicted in Fig.~\ref{fig:abc-set}.
The task is to write the letter `B' without any mistakes.
The purpose of this experiment is to reveal the characteristics of each method. 
For example, it remains unknown how many autoregressions are appropriate for each comparative method.
It is difficult to provide an answer directly because the more the autoregressive number increases, the more training time is needed. 
Hence, it is required to know model characteristics and utilize them for subsequent experiments.

\subsubsection{Data Collection \& Training}
In total, 18 motion data were collected with bilateral control at three paper heights: 70, 45, and 20 mm. 
The task duration was approximately 6.5 seconds (325 time-steps) and that of each data was 28.02 seconds (1401 time-steps). Thus, each dataset contains four demonstrations.
In training, the parameters are determined by Bayesian optimization with optuna\cite{optuna}, as presented in Table~\ref{params-exp1}.

\subsubsection{Task Validation}
After the training of each model, the autonomous operation was executed at five paper heights of 70, 60, 45, 35, and 20 mm.
Note that the paper heights 60 and 35 mm were not trained. 
The standard of the task's success is determined considering whether the follower succeeded in writing the letter `B' without any mistakes five times sequentially.

\subsubsection{Experiment results}
As presented in Table \ref{result-exp1}, autonomous operation tends to be unsuccessful when the autoregressive number is low. 
This result is as predicted because previous research has shown that autoregressive learning is effective in improving the performance of autonomous operations \cite{sm2sm,s2sm}. 
Therefore, the autoregressive number was fixed at 20 in the comparative methods during subsequent experiments.
However, the training of the FL2FL model with MTRNN could not be completed within 24 h. Thus, its autoregressive number was fixed at 1.

% \usepackage{multirow}

\begin{table*}[]

\caption{Hyper parameters gained by optuna in experiment 1. These were calculated from the bayesian optimization by optuna. In the column of the CW-RNN, ``Node'' represents the node size, and ``module'' means the number of divisions.
Also in other columns,`` Node'' means the node size.
In the column of MTRNN, ``$\tau$'' means the time constant.
In the columns of LSTM, ``LSTM Cells'' represents the number of the stacked LSTM cells.
A more detailed explanation of these hyperparameters can be found in the section \ref{sec:system}.}
\label{params-exp1}
\begin{tabular}{lllllllll}
\cline{1-8}
\multicolumn{2}{|l|}{CW-RNN}                                                                                                                                         & \multicolumn{2}{l|}{FS-RNN}                                   & \multicolumn{2}{l|}{MTRNN}                                                                                                                                                                                                                                & \multicolumn{2}{l|}{LSTM}                                                                                                                                         &  \\ \cline{1-8}
\multicolumn{1}{|l|}{F2FL}                                                        & \multicolumn{1}{l|}{FL2FL}                                                       & \multicolumn{1}{l|}{F2FL}     & \multicolumn{1}{l|}{FL2FL}    & \multicolumn{1}{l|}{F2L}                                                                                                    & \multicolumn{1}{l|}{FL2FL}                                                                                                  & \multicolumn{1}{l|}{F2FL}                                                       & \multicolumn{1}{l|}{FL2FL}                                                      &  \\ \cline{1-8}
\multicolumn{1}{|l|}{\begin{tabular}[c]{@{}l@{}}module:5\\ Node:400\end{tabular}} & \multicolumn{1}{l|}{\begin{tabular}[c]{@{}l@{}}module:5\\ Node:400\end{tabular}} & \multicolumn{1}{l|}{Node:650} & \multicolumn{1}{l|}{Node:650} & \multicolumn{1}{l|}{\begin{tabular}[c]{@{}l@{}}Node\_fast:550\\ Node\_slow:150\\ $\tau$\_fast:10\\ $\tau$\_slow:350\end{tabular}} & \multicolumn{1}{l|}{\begin{tabular}[c]{@{}l@{}}Node\_fast:450\\ Node\_slow:250\\ $\tau$\_fast:40\\ $\tau$\_slow:100\end{tabular}} & \multicolumn{1}{l|}{\begin{tabular}[c]{@{}l@{}}Node:200\\ LSTM\\Cells:2\end{tabular}} & \multicolumn{1}{l|}{\begin{tabular}[c]{@{}l@{}}Node:180\\ LSTM\\Cells:1\end{tabular}} &  \\ \cline{1-8}
\multicolumn{2}{l}{}                                                                                                                                                 &                               &                               &                                                                                                                             &                                                                                                                             &                                                                                 &                                                                                 &  \\
                                                                                  &                                                                                  &                               &                               &                                                                                                                             &                                                                                                                             &                                                                                 &                                                                                 &  \\
                                                                                  &                                                                                  &                               &                               &                                                                                                                             &                                                                                                                             &                                                                                 &                                                                                 & 
\end{tabular}
\end{table*}

\begin{table*}[]
\caption{Result of experiment 1 (Writing the letter B). Each check mark means success at all five heights.\\Because the training of the FL2FL model with MTRNN was not finished within 24 hours, their columns say "None".}
\label{result-exp1}
\begin{tabular}{|l|ll|ll|ll|ll|}
\hline
                                                                & \multicolumn{2}{l|}{CW-RNN}                & \multicolumn{2}{l|}{FS-RNN}            & \multicolumn{2}{l|}{MTRNN}                              & \multicolumn{2}{l|}{LSTM}                  \\ \hline
\begin{tabular}[c]{@{}l@{}}Autoregressive\\ number\end{tabular} & \multicolumn{1}{l|}{F2FL}      & FL2FL     & \multicolumn{1}{l|}{F2FL}      & FL2FL & \multicolumn{1}{l|}{F2L}                        & FL2FL & \multicolumn{1}{l|}{F2FL}      & FL2FL     \\ \hline
1 (w/o autoregression)                                                               & \multicolumn{1}{l|}{-}         & -         & \multicolumn{1}{l|}{-}         & -     & \multicolumn{1}{l|}{\multirow{3}{*}{\checkmark}} & -     & \multicolumn{1}{l|}{-}         & -         \\ \cline{1-5} \cline{7-9} 
5                                                               & \multicolumn{1}{l|}{-}         & -         & \multicolumn{1}{l|}{-}         & -     & \multicolumn{1}{l|}{}                           & None  & \multicolumn{1}{l|}{-}         & -         \\ \cline{1-5} \cline{7-9} 
20                                                              & \multicolumn{1}{l|}{\checkmark} & \checkmark & \multicolumn{1}{l|}{\checkmark} & -     & \multicolumn{1}{l|}{}                           & None  & \multicolumn{1}{l|}{\checkmark} & \checkmark \\ \hline
\end{tabular}
\end{table*}

\subsubsection{Evaluation of the autoregressive learning with the PE divergence}
In the previous section, the effect of the autoregressive learning was qualitatively given. Then, its quantitative performance is analyzed here by comparing the PE divergence of the LSTM model (our conventional model) with and without the autoregressive learning.
In this analysis, the PE divergence between the autonomous operation and training data was computed.
In Fig.~\ref{fig:auto-ope}, the follower response value during the autonomous operation and state in the training data are plotted. 
In addition, a Hankel matrix was created from this time series, and the PE divergence was calculated as depicted in Fig.~\ref{fig:auto-pe}.
As a result, the PE divergence was high in the case of operation without autoregressive learning.
This result proves that autoregressive learning reduces the influence of the covariate shift and helps robots perform tasks successfully.

\Figure[t!](topskip=0pt, botskip=0pt, midskip=0pt)[width=6cm]{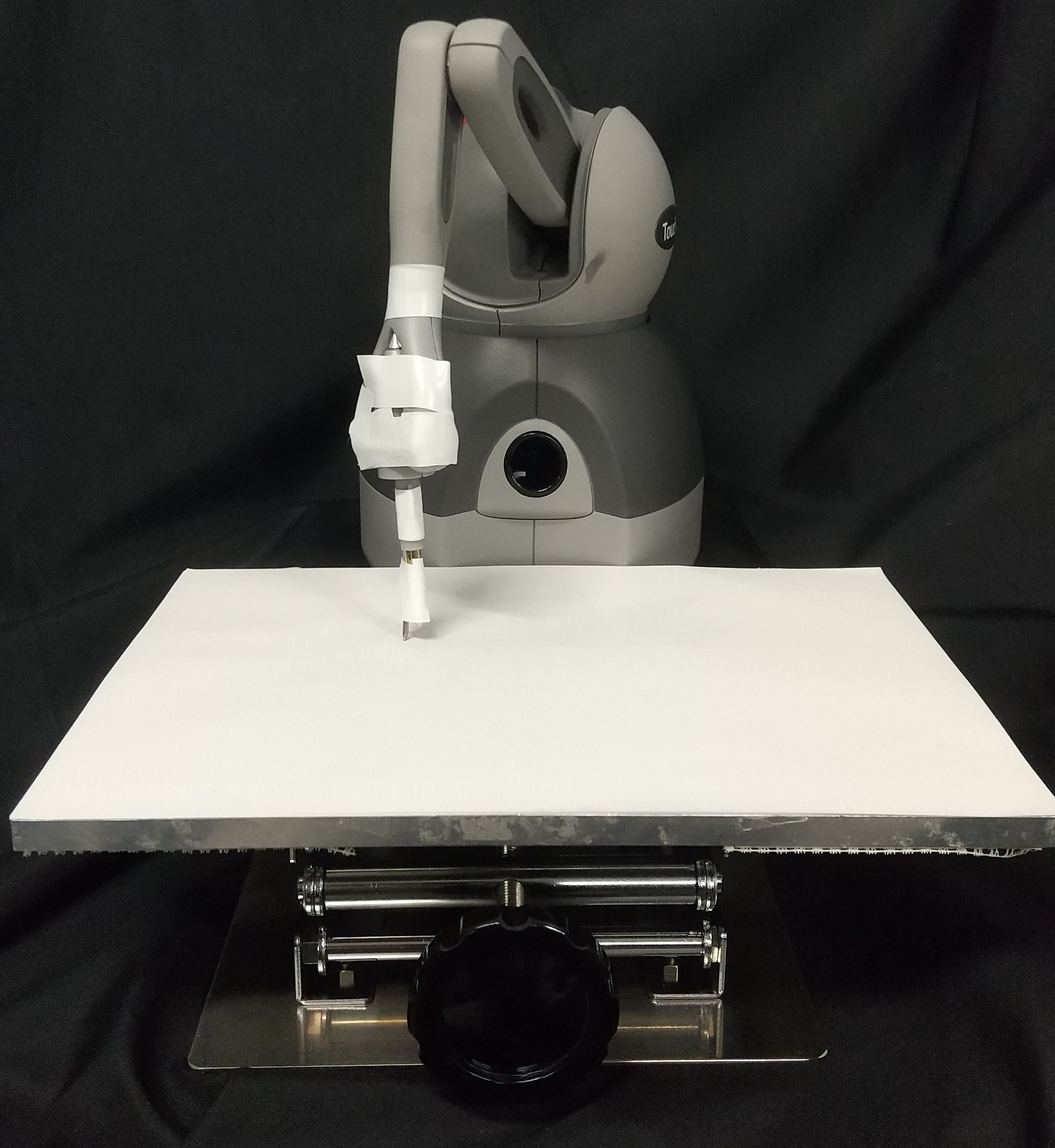}
{The settings of experiments 1, 2, and 3.\label{fig:abc-set}}

\Figure[t!](topskip=0pt, botskip=0pt, midskip=0pt)[width=18cm]{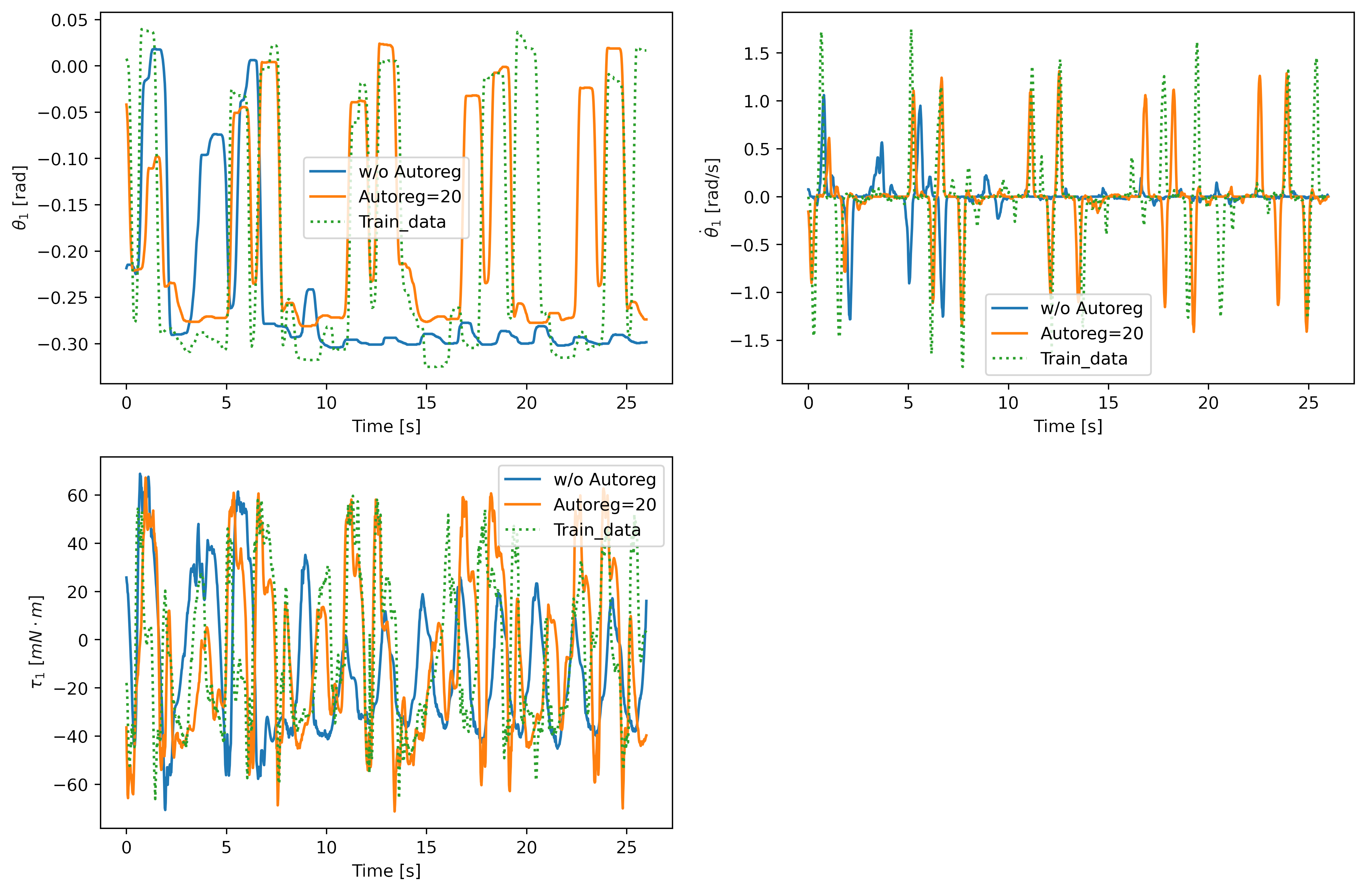}
{\label{fig:auto-ope}Autonomous operation with LSTM and training data in experiment 1 when the paper height was 20 mm. From the left, the first angle, angular velocity, and torque are shown. }

\Figure[t!](topskip=0pt, botskip=0pt, midskip=0pt)[width=18cm]{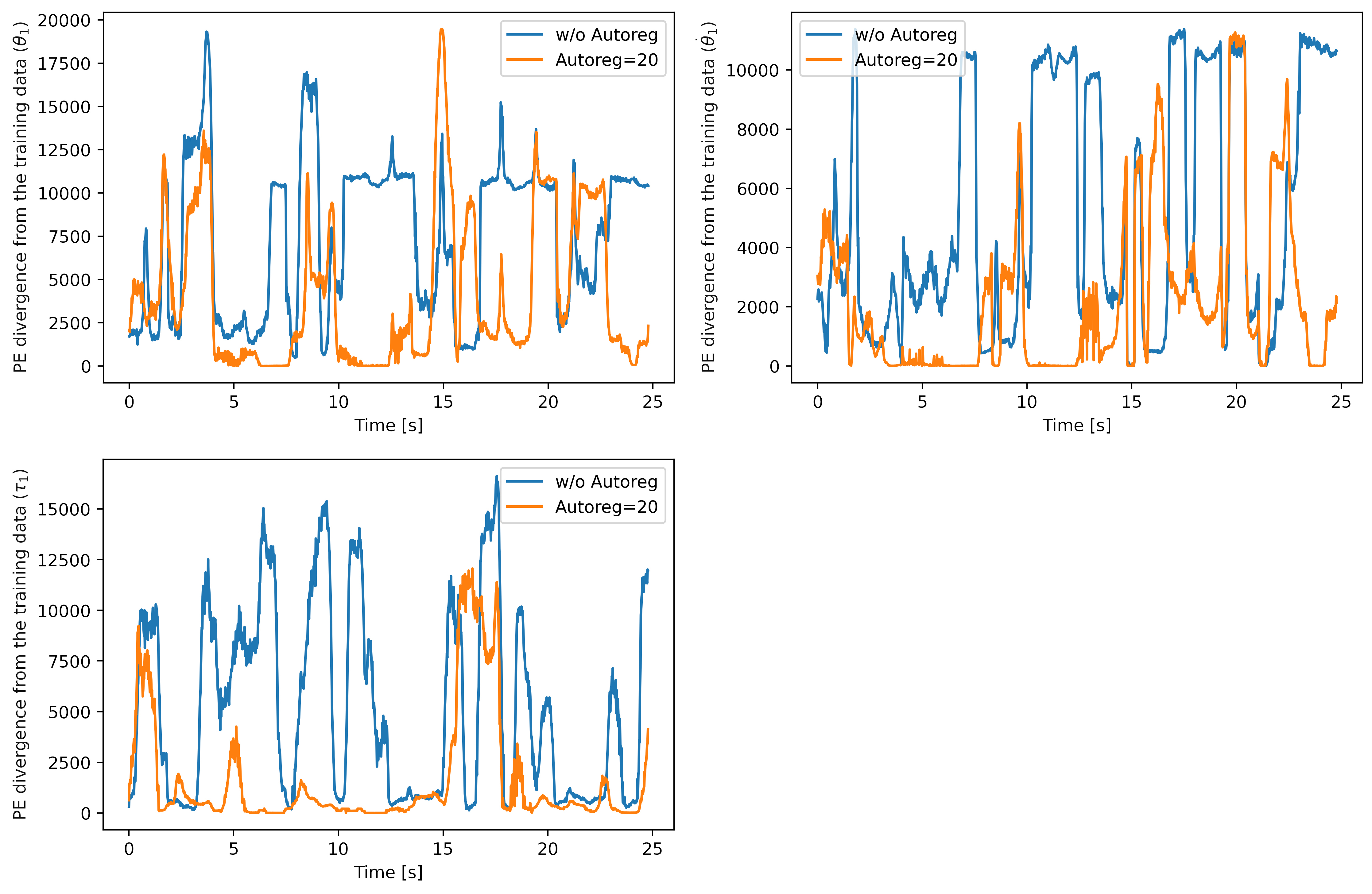}
{\label{fig:auto-pe}PE divergence between autonomous operation with LSTM and training data in experiment 1 when the paper height was 20 mm. 
From the left, the first angle, angular velocity, and torque are shown.}

\subsection{Experiment 2 (Writing the characters ABC)}
\label{writeABC}
\subsubsection{Task Design}
The goal of this task was to write three letters, ``A,'' ``B,'' and ``C'' sequentially without any mistakes.
The execution of this task takes approximately 20 s (1000 steps). In addition, adaptation to changes in height is necessary.
Therefore, the robot needs to be able to both plan long-term tasks and adjust the force properly.

\subsubsection{Data Collection \& Training}
The motion data were collected with bilateral control. Here, demonstrations were conducted at paper heights of 70 mm, 45 mm, and 20 mm. A total of 30 data points were collected.
Moreover, the duration of each data is 60.4 s (3020 time-steps) and that of the task is approximately 20 s (1000 time-steps). Thus, each dataset contained three demonstrations.
As shown in Fig. \ref{fig:abc-paper}, experts wrote along the printed letters. In training, hyperparameters are determined according to a Bayesian optimization with optuna\cite{optuna}, as reported in Table \ref{param-prop-2} and \ref{param-comp-2}.

\subsubsection{Task Validation}
After training each model, the autonomous operation was executed at five paper heights of 70, 60, 45, 35, and 20 mm.
Note that the paper heights 60 and 35 mm were not trained. 
First, we focused on the epoch and loss graphs and identified an epoch number $E_{med}$ that could perform the task effectively. Subsequently, we performed autonomous operations with five models with epoch numbers of $E_{med}, E_{med}\pm1000, E_{med}\pm2000$, and calculated their success rates.
In the autonomous operation using the proposed methods, this procedure was performed on the upper layer.
During the training of the proposed methods, the autoregressive numbers of the upper and lower layers were 1, 5, and 20, respectively. As a result, the upper layer of the autoregressive number needed to be 5 or 20 to successfully execute the tasks.  However, the lower layer did not require autoregression. Therefore, the autoregressive number of the upper layer was five and that of the lower layer was one in this experiment because small autoregressive numbers are preferable to shorten the training time.

\subsubsection{Experimental Result}
Table \ref{success-abc} reports the success rates of the experiment. In the row ``Success or Failure'', ``Success''  denotes that a model that can execute tasks at every height was found.
``Partial Success'' denotes that a model that can execute tasks at limited heights was found. ``Failure'' denotes that a model that can execute tasks was not found.
Our proposed methods and CW-RNN and MTRNN succeeded in performing tasks at every height. 
Among these methods, the second proposed method type exhibited a higher success rate and reproducible trajectories, as depicted in Fig.~\ref{fig:abc-trajectory} and Table \ref{success-abc}.
During autonomous operation, a misalignment of the fixed ballpoint pen sometimes occurs. This phenomenon had a negative influence on torque information. However, its influence was limited in the second proposed method because the upper layer only included positional (angular) information. 

On the other hand, the first proposed method required the least time to train, as reported in Table \ref{train-abc}. Both proposed methods require 1/20 less training time compared to the other methods. This is because their upper layer timescales were large, and their input and output shapes were symmetric.

% \usepackage{multirow}

\begin{table*}[]
\caption{Hyper parameters of the proposed methods in experiment 2. These hyperparameters were calculated from the bayesian optimization by optuna. ``Node'' represents the node size, and  ``LSTM Cells'' represents the number of the stacked LSTM cells.
A more detailed explanation of these hyperparameters can be found in the section \ref{sec:system}.}
\label{param-prop-2}
\begin{tabular}{|l|ll|ll|}
\hline
\multirow{2}{*}{} & \multicolumn{2}{l|}{Prop. 1}                                                                                                                  & \multicolumn{2}{l|}{Prop. 2}                                                                                                                 \\ \cline{2-5} 
                  & \multicolumn{1}{l|}{F2FL}                                                       & FL2FL                                                      & \multicolumn{1}{l|}{F2FL}                                                       & FL2FL                                                     \\ \hline
Upper             & \multicolumn{2}{l|}{\begin{tabular}[c]{@{}l@{}}Node:120\\ LSTM\\Cells:1\end{tabular}}                                                              & \multicolumn{2}{l|}{\begin{tabular}[c]{@{}l@{}}Node:170\\ LSTM\\Cells:1\end{tabular}}                                                             \\ \hline
Lower             & \multicolumn{1}{l|}{\begin{tabular}[c]{@{}l@{}}Node:100\\ LSTM\\Cells:1\end{tabular}} & \begin{tabular}[c]{@{}l@{}}Node:120\\ LSTM\\Cells:1\end{tabular} & \multicolumn{1}{l|}{\begin{tabular}[c]{@{}l@{}}Node:170\\ LSTM\\Cells:1\end{tabular}} & \begin{tabular}[c]{@{}l@{}}Node:90\\ LSTM\\Cells:1\end{tabular} \\ \hline
\end{tabular}
\end{table*}

\begin{table*}[]
\caption{Hyper parameters of the comparative methods in experiment 2. These were calculated from the bayesian optimization by optuna. In the column of the CW-RNN, ``Node'' represents the node size, and ``module'' means the number of divisions.
Also in other columns,`` Node'' means the node size.
In the column of MTRNN, ``$\tau$'' means the time constant.
In the columns of LSTM, ``LSTM Cells'' represents the number of the stacked LSTM cells.
A more detailed explanation of these hyperparameters can be found in the section \ref{sec:system}.}
\label{param-comp-2}
\begin{tabular}{|ll|ll|ll|ll|}
\hline
\multicolumn{2}{|l|}{CW-RNN}                                                                                                                    & \multicolumn{2}{l|}{FS-RNN}              & \multicolumn{2}{l|}{MTRNN}                                                                                                                                                                                                            & \multicolumn{2}{l|}{LSTM}                                                                                                                   \\ \hline
\multicolumn{1}{|l|}{F2FL}                                                        & FL2FL                                                       & \multicolumn{1}{l|}{F2FL}     & FL2FL    & \multicolumn{1}{l|}{F2L}                                                                                                    & FL2FL                                                                                                   & \multicolumn{1}{l|}{F2FL}                                                       & FL2FL                                                     \\ \hline
\multicolumn{1}{|l|}{\begin{tabular}[c]{@{}l@{}}module:4\\ Node:480\end{tabular}} & \begin{tabular}[c]{@{}l@{}}module:5\\ Node:450\end{tabular} & \multicolumn{1}{l|}{Node:550} & Node:650 & \multicolumn{1}{l|}{\begin{tabular}[c]{@{}l@{}}Node\_fast:550\\ Node\_slow:250\\ $\tau$\_fast:10\\ $\tau$\_slow:150\end{tabular}} & \begin{tabular}[c]{@{}l@{}}Node\_fast:500\\ Node\_slow:400\\ $\tau$\_fast:190\\ $\tau$\_slow:250\end{tabular} & \multicolumn{1}{l|}{\begin{tabular}[c]{@{}l@{}}Node:110\\ LSTM\\Cells:1\end{tabular}} & \begin{tabular}[c]{@{}l@{}}Node:70\\ LSTM\\Cells:1\end{tabular} \\ \hline
\end{tabular}
\end{table*}

\begin{table*}[]
\caption{The autonomous operation success rate of every model in experiment 2.}
\label{success-abc}
\begin{tabular}{|l|ll|ll|ll|ll|ll|ll|}
\hline
\multirow{2}{*}{\begin{tabular}[c]{@{}l@{}}Paper\\ height {[}mm{]}\end{tabular}} & \multicolumn{2}{l|}{Prop. 1}          & \multicolumn{2}{l|}{\textbf{Prop. 2}}                                                              & \multicolumn{2}{l|}{CW-RNN}          & \multicolumn{2}{l|}{FS-RNN}                                                            & \multicolumn{2}{l|}{MTRNN}           & \multicolumn{2}{l|}{LSTM}         \\ \cline{2-13} 
                                                                                 & \multicolumn{1}{l|}{F2FL}    & FL2FL & \multicolumn{1}{l|}{\textbf{F2FL}}    & FL2FL                                                     & \multicolumn{1}{l|}{F2FL}    & FL2FL & \multicolumn{1}{l|}{F2FL}                                                      & FL2FL & \multicolumn{1}{l|}{F2L}     & FL2FL & \multicolumn{1}{l|}{F2FL} & FL2FL \\ \hline
70                                                                               & \multicolumn{1}{l|}{100\%}   & -     & \multicolumn{1}{l|}{\textbf{100\%}}   & 100\%                                                     & \multicolumn{1}{l|}{40\%}    & -     & \multicolumn{1}{l|}{80\%}                                                      & -     & \multicolumn{1}{l|}{40\%}    & -     & \multicolumn{1}{l|}{-}    & -     \\ \hline
60                                                                               & \multicolumn{1}{l|}{40\%}    & -     & \multicolumn{1}{l|}{\textbf{100\%}}   & 0\%                                                       & \multicolumn{1}{l|}{40\%}    & -     & \multicolumn{1}{l|}{40\%}                                                      & -     & \multicolumn{1}{l|}{40\%}    & -     & \multicolumn{1}{l|}{-}    & -     \\ \hline
45                                                                               & \multicolumn{1}{l|}{100\%}   & -     & \multicolumn{1}{l|}{\textbf{100\%}}   & 100\%                                                     & \multicolumn{1}{l|}{40\%}    & -     & \multicolumn{1}{l|}{60\%}                                                      & -     & \multicolumn{1}{l|}{60\%}    & -     & \multicolumn{1}{l|}{-}    & -     \\ \hline
35                                                                               & \multicolumn{1}{l|}{40\%}    & -     & \multicolumn{1}{l|}{\textbf{100\%}}   & 100\%                                                     & \multicolumn{1}{l|}{80\%}    & -     & \multicolumn{1}{l|}{20\%}                                                      & -     & \multicolumn{1}{l|}{80\%}    & -     & \multicolumn{1}{l|}{-}    & -     \\ \hline
20                                                                               & \multicolumn{1}{l|}{100\%}   & -     & \multicolumn{1}{l|}{\textbf{100\%}}   & 100\%                                                     & \multicolumn{1}{l|}{80\%}    & -     & \multicolumn{1}{l|}{20\%}                                                      & -     & \multicolumn{1}{l|}{80\%}    & -     & \multicolumn{1}{l|}{-}    & -     \\ \hline
\begin{tabular}[c]{@{}l@{}}Success\\ or\\ Failure\end{tabular}                      & \multicolumn{1}{l|}{Success} & Failure  & \multicolumn{1}{l|}{\textbf{Success}} & \begin{tabular}[c]{@{}l@{}}Partial\\ Success\end{tabular} & \multicolumn{1}{l|}{Success} & Failure  & \multicolumn{1}{l|}{\begin{tabular}[c]{@{}l@{}}Partial\\ Success\end{tabular}} & Failure  & \multicolumn{1}{l|}{Success} & Failure  & \multicolumn{1}{l|}{Failure} & Failure  \\ \hline
\end{tabular}
\end{table*}

\begin{table*}[]
\caption{Training time of each model in experiment 2 (writing ABC).}
\label{train-abc}
\begin{tabular}{|ll|ll|ll|llllllll|}
\hline
\multicolumn{2}{|l|}{\multirow{2}{*}{}}                    & \multicolumn{2}{l|}{\textbf{Prop. 1}}        & \multicolumn{2}{l|}{Prop. 2}        & \multicolumn{2}{l|}{CW-RNN}                             & \multicolumn{2}{l|}{FS-RNN}                            & \multicolumn{2}{l|}{MTRNN}                              & \multicolumn{2}{l|}{LSTM}         \\ \cline{3-14} 
\multicolumn{2}{|l|}{}                                     & \multicolumn{1}{l|}{\textbf{F2FL}}  & FL2FL & \multicolumn{1}{l|}{F2FL}  & FL2FL & \multicolumn{1}{l|}{F2FL}  & \multicolumn{1}{l|}{FL2FL} & \multicolumn{1}{l|}{F2FL} & \multicolumn{1}{l|}{FL2FL} & \multicolumn{1}{l|}{F2L}   & \multicolumn{1}{l|}{FL2FL} & \multicolumn{1}{l|}{F2FL} & FL2FL \\ \hline
\multicolumn{1}{|l|}{\multirow{2}{*}{Upper}} & Epoch       & \multicolumn{1}{l|}{\textbf{40000}} & -     & \multicolumn{1}{l|}{26000} & -     & \multicolumn{8}{l|}{\multirow{2}{*}{}}                                                                                                                                                                         \\ \cline{2-6}
\multicolumn{1}{|l|}{}                       & Time{[}s{]} & \multicolumn{1}{l|}{\textbf{1300}}  & -     & \multicolumn{1}{l|}{1815}  & -     & \multicolumn{8}{l|}{}                                                                                                                                                                                          \\ \hline
\multicolumn{1}{|l|}{\multirow{2}{*}{Lower}} & Epoch       & \multicolumn{1}{l|}{\textbf{2400}}  & -     & \multicolumn{1}{l|}{1000}  & -     & \multicolumn{1}{l|}{28000} & \multicolumn{1}{l|}{-}     & \multicolumn{1}{l|}{-}    & \multicolumn{1}{l|}{-}     & \multicolumn{1}{l|}{92000} & \multicolumn{1}{l|}{-}     & \multicolumn{1}{l|}{-}    & -     \\ \cline{2-14} 
\multicolumn{1}{|l|}{}                       & Time{[}s{]} & \multicolumn{1}{l|}{\textbf{898}}   & -     & \multicolumn{1}{l|}{183}   & -     & \multicolumn{1}{l|}{37816} & \multicolumn{1}{l|}{-}     & \multicolumn{1}{l|}{-}    & \multicolumn{1}{l|}{-}     & \multicolumn{1}{l|}{71024} & \multicolumn{1}{l|}{-}     & \multicolumn{1}{l|}{-}    & -     \\ \hline
\end{tabular}
\end{table*}

\Figure[t!](topskip=0pt, botskip=0pt, midskip=0pt)[width=8cm]{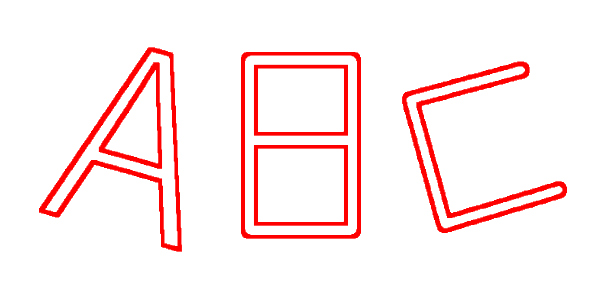}
{The letters printed on the paper for experiment 2. \label{fig:abc-paper}}

\Figure[t!](topskip=0pt, botskip=0pt, midskip=0pt)[width=8cm]{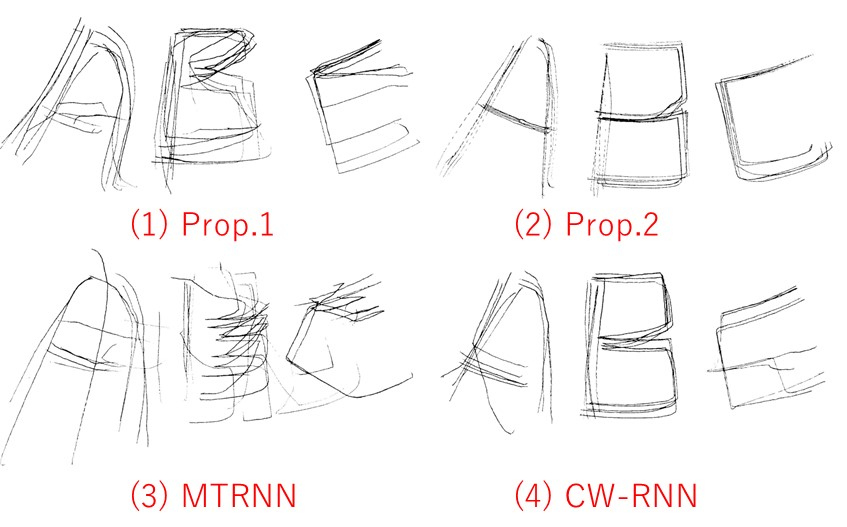}
{Letters written by each method.\label{fig:abc-trajectory}}

\subsection{Experiment 3 (Writing unlearned characters)}
\subsubsection{Task Design}
In this experiment, the most notable advantage of our proposed method was proven.
The task was to write three letters, “X,” “Y,” and “Z” sequentially without any mistakes. 
In our methods, the upper and lower layers were independently trained.
During the training, the upper layer learns the long-term task plan, while the lower layer learns the motion primitive.
Hence, various untrained tasks can be executed by utilizing the trained lower layer.
Furthermore, this approach improves computational efficiency.

Especially in proposed method II, the upper layer includes only angle information and not force information. The positional (angular) trajectories can be computed from the kinematics or obtained by direct teaching. In another way, the higher layer can be trained using reinforcement-learning-based methods, such as sim2real\cite{sim1}.
Therefore, rough positional trajectories can be created using such methods, and the positional, velocity, and force commands can be generated by the lower layer. Using these methods, various tasks can be executed.
In this experiment, we used the lower layer of the proposed method II trained in experiment 2, and the collected motion data were input to the $[\hat{\theta}_{t+10},\hat{\theta}_{t+20},\hat{\theta}_{t+30}, \hat{\theta}_{t+40}]$ parameters of the lower layer.

\subsubsection{Data Collection}
In the data collection phase, the follower robot was manipulated like a direct teaching approach, and the angular response of the follower robot was recorded.
At this time, data collection was conducted to write along frames printed on the paper, as presented in Fig. \ref{fig:xyz-paper}.
Note that bilateral control was not used, and training was not conducted here.
\subsubsection{Task Validation}
Subsequently, the recorded follower angular responses were input to the $[\hat{\theta}_{t+10},\hat{\theta}_{t+20},\hat{\theta}_{t+30}, \hat{\theta}_{t+40}]$ parameters of the lower layer of the second proposed method.

\subsubsection{Experimental results}
As depicted in Fig. \ref{fig:xyz-paper2}, the outline shape was successfully written by exchanging the upper layer for the angular trajectory collected in advance. 
%This result indicates that the trained lower layer is helpful for executing various unlearned tasks.
This result indicates that the trained lower layer appropriately generated angle, angular velocity and torque commands only from angular responses and is helpful for executing various unlearned tasks.
However, its shape was slightly different from that of the recorded follower state because of the limited diversity of training data.

\Figure[t!](topskip=0pt, botskip=0pt, midskip=0pt)[width=4cm]{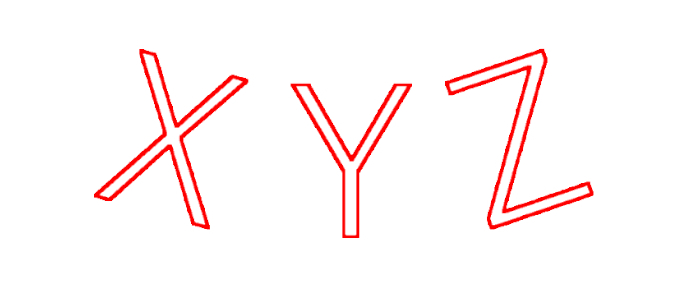}
{The letters printed on the paper for experiment 3. \label{fig:xyz-paper}}

\Figure[t!](topskip=0pt, botskip=0pt, midskip=0pt)[width=4cm]{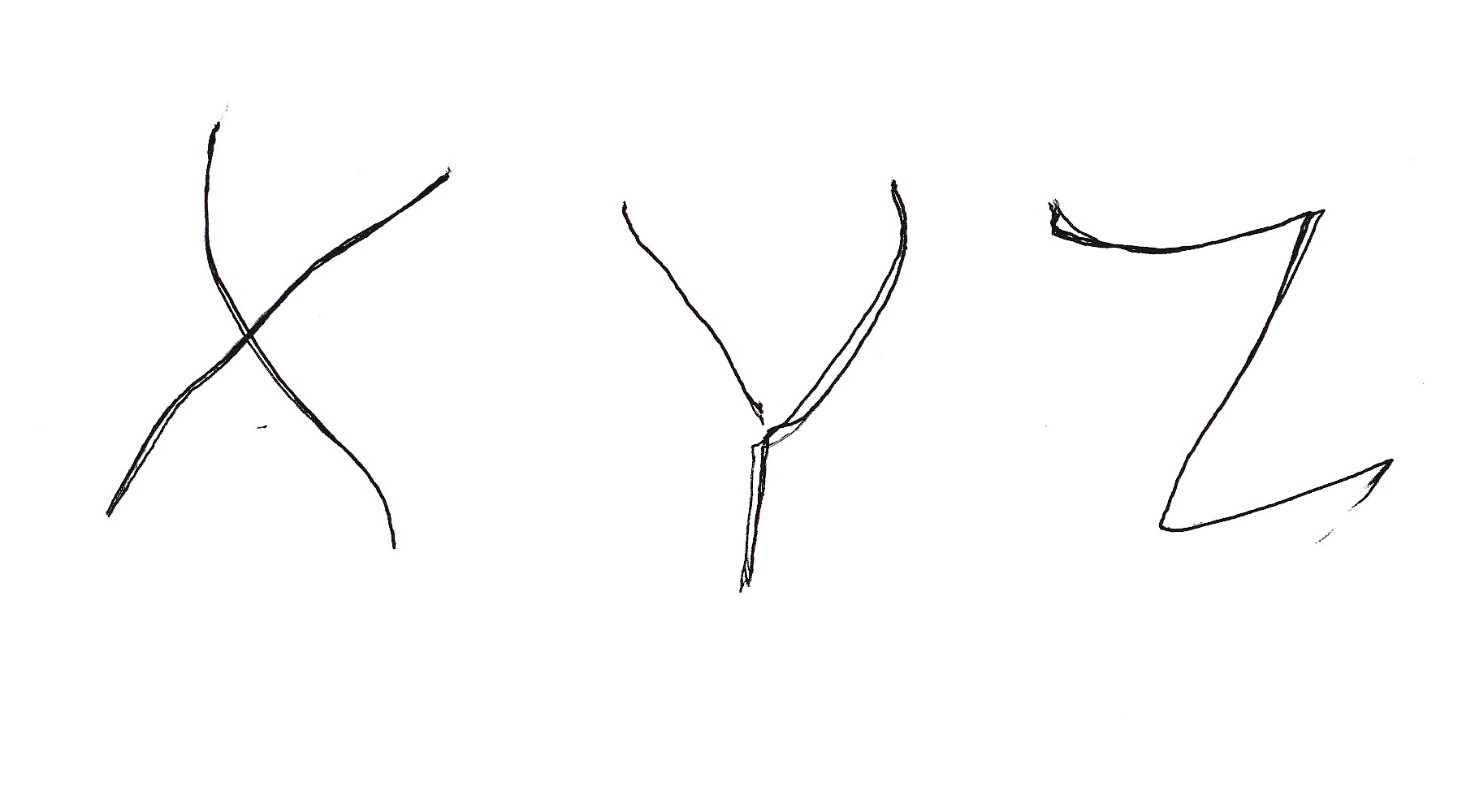}
{Letters written in experiment 3. \label{fig:xyz-paper2}}

\subsection{Experiment 4 (Wiping the whiteboard with the eraser)}
\subsubsection{Task Design}
Figs. \ref{fig:wipe-set} and \ref{fig:wipe-set2} present the setting of this experiment.
In this experiment, the whiteboard eraser was fixed to the follower robot. The task was to wipe the tilted whiteboard in three different directions.
Executing this task approximately required 30 s (1500 steps). Additionally, the follower robot was required to adapt to the changes in the tilt angle of the whiteboard. 
Therefore, the robot needed to be able to both plan long-term tasks and adjust the force properly.
Moreover, note that the fixed whiteboard eraser periodically moved from side to side while wiping.
This means that our approach is also effective for other tasks, such as grinding off parts.

\Figure[t!](topskip=0pt, botskip=0pt, midskip=0pt)[width=8cm]{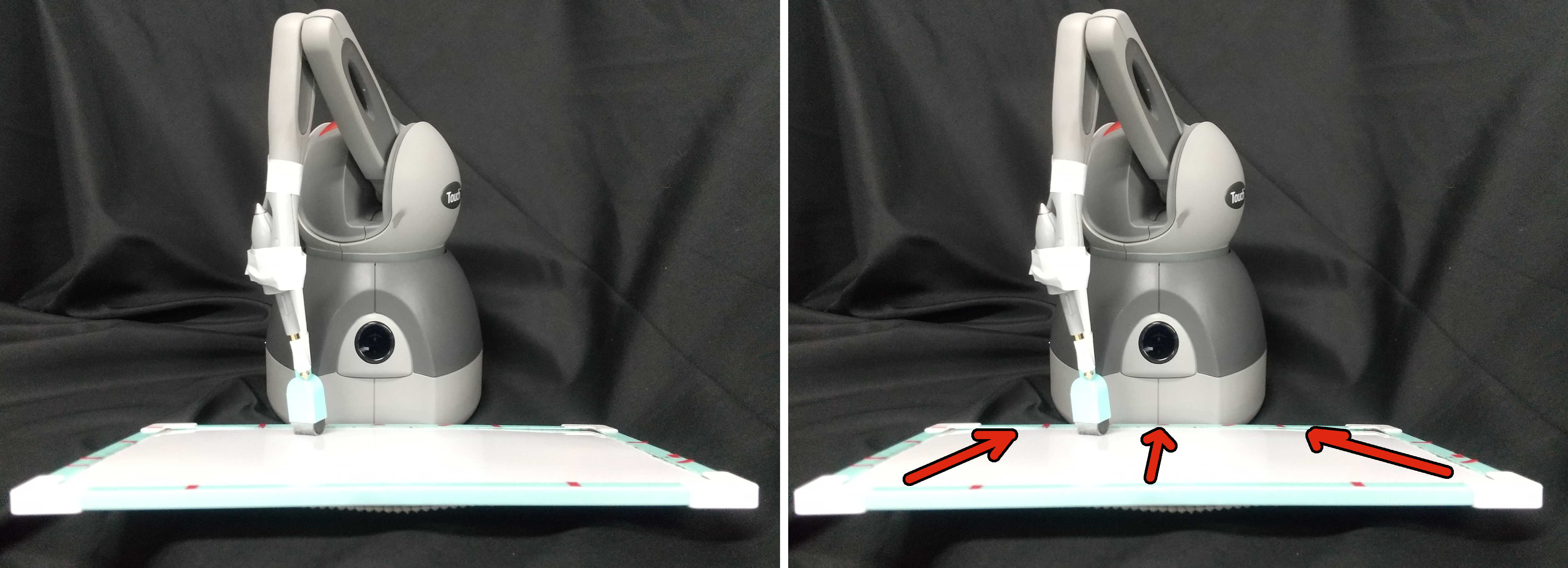}
{The left figure presents the setting of experiment 4, and the right figure shows three erasing directions.\label{fig:wipe-set}}

\Figure[t!](topskip=0pt, botskip=0pt, midskip=0pt)[width=6cm]{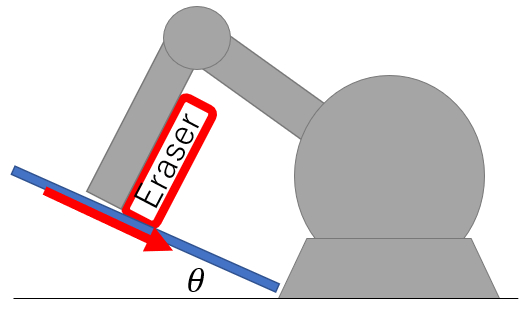}
{The side view of the setting of experiment 4.\label{fig:wipe-set2}}

\subsubsection{Data Collection \& Training}
In this task, a robot wiped the tilted plate in three directions.
Here, the robot rubbed the plate with a fixed eraser, similar to using an eraser manually.
In total, 9 data were collected on a tilted plate with bilateral control.
In these demonstrations, the tilt angles of the plate were 0°, 10°, and 20 °. 
Furthermore, the data duration was 60.40 s (3020 timesteps) and that of the task was approximately 30 s (1500 time-steps). Therefore, each data point contains two demonstrations.
In addition to the second experiment, in training, the hyperparameters were determined by Bayesian optimization with optuna\cite{optuna}, as reported in Tables \ref{param-prop-4} and \ref{param-comp-4}.

\subsubsection{Task Validation}
After training each model, the autonomous operation was executed at five tilt angles of: 20°, 15°, 10°, 5°, and 0°.
Note that the tilt angles 15° and 5° were not trained. 
First, we focused on the epoch and loss graphs and identified an epoch number $E_{med}$ that could perform the task. Subsequently, we performed autonomous operations with five models with epoch numbers of $E_{med}, E_{med}\pm1000, E_{med}\pm2000$, and calculated the model success rates. In the autonomous operation with the proposed methods, this procedure was performed on the upper layer.
When training the proposed methods, the autoregressive numbers of the upper and lower layers were 1, 5, and 20, respectively. As a result, the upper and lower layer autoregression numbers needed to be 5 or 20 to execute the tasks successfully.  Therefore, the autoregression number of both layers was fixed at five.
In this task, a success criterion is whether a robot can wipe in all three directions without mistakes.

\subsubsection{Experimental result}
The success rate of the experiment is reported in Table \ref{success-wipe}. In the row ``Success or Failure,'' ``Success''  denotes that a model that can execute tasks at every tilt angle was found. ``Partial Success'' denotes that a model that can execute tasks at certain tilt angles was found. ``Failure'' denotes that a model that can execute tasks was not found.
Our proposed methods and MTRNN succeeded in performing tasks at every angle, as reported in Table \ref{train-wipe}. 
In these methods, the first proposed method type exhibited a higher success rate. 
This is because the upper layer of the first proposed method includes the force information, which helps the robot recognize the current situation in a long-term task. In addition, both proposed methods required 1/20 of the training time required by the comparative methods.

\begin{table*}[]
\caption{Hyper parameters of the proposed method in experiment 4. These hyperparameters were calculated from the bayesian optimization by optuna. ``Node'' represents the node size, and  ``LSTM Cells'' represents the number of the stacked LSTM cells.
A more detailed explanation of these hyperparameters can be found in the section \ref{sec:system}.}
\label{param-prop-4}
\begin{tabular}{|l|ll|ll|}
\hline
\multirow{2}{*}{} & \multicolumn{2}{l|}{Prop. 1}                                                                                                                  & \multicolumn{2}{l|}{Prop. 2}                                                                                                                  \\ \cline{2-5} 
                  & \multicolumn{1}{l|}{F2FL}                                                       & FL2FL                                                      & \multicolumn{1}{l|}{F2FL}                                                       & FL2FL                                                      \\ \hline
Upper             & \multicolumn{2}{l|}{\begin{tabular}[c]{@{}l@{}}Node:120\\ LSTM\\Cells:1\end{tabular}}                                                              & \multicolumn{2}{l|}{\begin{tabular}[c]{@{}l@{}}Node:170\\ LSTM\\Cells:1\end{tabular}}                                                              \\ \hline
Lower             & \multicolumn{1}{l|}{\begin{tabular}[c]{@{}l@{}}Node:100\\ LSTM\\Cells:1\end{tabular}} & \begin{tabular}[c]{@{}l@{}}Node:160\\ LSTM\\Cells:1\end{tabular} & \multicolumn{1}{l|}{\begin{tabular}[c]{@{}l@{}}Node:170\\ LSTM\\Cells:1\end{tabular}} & \begin{tabular}[c]{@{}l@{}}Node:170\\ LSTM\\Cells:1\end{tabular} \\ \hline
\end{tabular}
\end{table*}

\begin{table*}[]
\caption{Hyper parameters of the comparative methods in experiment 4. These were calculated from the bayesian optimization by optuna. In the column of the CW-RNN, ``Node'' represents the node size, and ``module'' means the number of divisions.
Also in other columns,`` Node'' means the node size.
In the column of MTRNN, ``$\tau$'' means the time constant.
In the columns of LSTM, ``LSTM Cells'' represents the number of the stacked LSTM cells.
A more detailed explanation of these hyperparameters can be found in the section \ref{sec:system}.}
\label{param-comp-4}
\begin{tabular}{|ll|ll|ll|ll|}
\hline
\multicolumn{2}{|l|}{CW-RNN}                                                                                                                     & \multicolumn{2}{l|}{FS-RNN}               & \multicolumn{2}{l|}{MTRNN}                                                                                                                                                                                                          & \multicolumn{2}{l|}{LSTM}                                                                                                                    \\ \hline
\multicolumn{1}{|l|}{F2FL}                                                         & FL2FL                                                       & \multicolumn{1}{l|}{F2FL}     & FL2FL     & \multicolumn{1}{l|}{F2L}                                                                                                   & FL2FL                                                                                                  & \multicolumn{1}{l|}{F2FL}                                                       & FL2FL                                                      \\ \hline
\multicolumn{1}{|l|}{\begin{tabular}[c]{@{}l@{}}module:4 \\ Node:360\end{tabular}} & \begin{tabular}[c]{@{}l@{}}module:3\\ Node:270\end{tabular} & \multicolumn{1}{l|}{Node:850} & Node:1150 & \multicolumn{1}{l|}{\begin{tabular}[c]{@{}l@{}}Node\_fast:550\\ Node\_slow:50\\ $\tau$\_fast:10\\ $\tau$\_slow:300\end{tabular}} & \begin{tabular}[c]{@{}l@{}}Node\_fast:450\\ Node\_slow:550\\ $\tau$\_fast:10\\ $\tau$\_slow:200\end{tabular} & \multicolumn{1}{l|}{\begin{tabular}[c]{@{}l@{}}Node:200\\ LSTM\\Cells:2\end{tabular}} & \begin{tabular}[c]{@{}l@{}}Node:130\\ LSTM\\Cells:1\end{tabular} \\ \hline
\end{tabular}
\end{table*}

\begin{table*}[]
\caption{The autonomous operation success rate of every model in experiment 4.}
\label{success-wipe}
\begin{tabular}{|l|ll|ll|ll|ll|ll|ll|}
\hline
\multirow{2}{*}{\begin{tabular}[c]{@{}l@{}}Degree of \\ a tilted plate {[}deg{]}\end{tabular}} & \multicolumn{2}{l|}{\textbf{Prop. 1}}          & \multicolumn{2}{l|}{Prop. 2}                                                              & \multicolumn{2}{l|}{CW-RNN}                                                            & \multicolumn{2}{l|}{FS-RNN}       & \multicolumn{2}{l|}{\textbf{MTRNN}}           & \multicolumn{2}{l|}{LSTM}         \\ \cline{2-13} 
                                                                                               & \multicolumn{1}{l|}{\textbf{F2FL}}    & FL2FL & \multicolumn{1}{l|}{F2FL}    & FL2FL                                                     & \multicolumn{1}{l|}{F2FL}                                                      & FL2FL & \multicolumn{1}{l|}{F2FL} & FL2FL & \multicolumn{1}{l|}{\textbf{S2M}}     & FL2FL & \multicolumn{1}{l|}{F2FL} & FL2FL \\ \hline
20                                                                                             & \multicolumn{1}{l|}{\textbf{100\%}}   & -     & \multicolumn{1}{l|}{100\%}   & 0\%                                                       & \multicolumn{1}{l|}{40\%}                                                      & -     & \multicolumn{1}{l|}{-}    & -     & \multicolumn{1}{l|}{\textbf{100\%}}   & -     & \multicolumn{1}{l|}{-}    & -     \\ \hline
15                                                                                             & \multicolumn{1}{l|}{\textbf{100\%}}   & -     & \multicolumn{1}{l|}{100\%}   & 0\%                                                       & \multicolumn{1}{l|}{60\%}                                                      & -     & \multicolumn{1}{l|}{-}    & -     & \multicolumn{1}{l|}{\textbf{80\%}}    & -     & \multicolumn{1}{l|}{-}    & -     \\ \hline
10                                                                                             & \multicolumn{1}{l|}{\textbf{80\%}}    & -     & \multicolumn{1}{l|}{80\%}    & 20\%                                                      & \multicolumn{1}{l|}{0\%}                                                       & -     & \multicolumn{1}{l|}{-}    & -     & \multicolumn{1}{l|}{\textbf{100\%}}   & -     & \multicolumn{1}{l|}{-}    & -     \\ \hline
5                                                                                              & \multicolumn{1}{l|}{\textbf{80\%}}    & -     & \multicolumn{1}{l|}{80\%}    & 0\%                                                       & \multicolumn{1}{l|}{40\%}                                                      & -     & \multicolumn{1}{l|}{-}    & -     & \multicolumn{1}{l|}{\textbf{80\%}}    & -     & \multicolumn{1}{l|}{-}    & -     \\ \hline
0                                                                                              & \multicolumn{1}{l|}{\textbf{100\%}}   & -     & \multicolumn{1}{l|}{60\%}    & 80\%                                                      & \multicolumn{1}{l|}{80\%}                                                      & -     & \multicolumn{1}{l|}{-}    & -     & \multicolumn{1}{l|}{\textbf{100\%}}   & -     & \multicolumn{1}{l|}{-}    & -     \\ \hline
\begin{tabular}[c]{@{}l@{}}Success \\ or \\ Failure\end{tabular}                                  & \multicolumn{1}{l|}{\textbf{Success}} & Failure  & \multicolumn{1}{l|}{Success} & \begin{tabular}[c]{@{}l@{}}Parital\\ Success\end{tabular} & \multicolumn{1}{l|}{\begin{tabular}[c]{@{}l@{}}Partial\\ Success\end{tabular}} & Failure  & \multicolumn{1}{l|}{Failure} & Failure  & \multicolumn{1}{l|}{\textbf{Success}} & Failure  & \multicolumn{1}{l|}{Failure} & Failure  \\ \hline
\end{tabular}
\end{table*}

\begin{table*}[]
\caption{Training time required by each model in experiment 4.}
\label{train-wipe}
\begin{tabular}{|ll|ll|ll|llllllll|}
\hline
\multicolumn{2}{|l|}{\multirow{2}{*}{}}                                                                    & \multicolumn{2}{l|}{\textbf{Prop. 1}}        & \multicolumn{2}{l|}{Prop. 2}        & \multicolumn{2}{l|}{CW-RNN}                            & \multicolumn{2}{l|}{FS-RNN}                            & \multicolumn{2}{l|}{MTRNN}                              & \multicolumn{2}{l|}{LSTM}         \\ \cline{3-14} 
\multicolumn{2}{|l|}{}                                                                                     & \multicolumn{1}{l|}{\textbf{F2FL}}  & FL2FL & \multicolumn{1}{l|}{F2FL}  & FL2FL & \multicolumn{1}{l|}{F2FL} & \multicolumn{1}{l|}{FL2FL} & \multicolumn{1}{l|}{F2FL} & \multicolumn{1}{l|}{FL2FL} & \multicolumn{1}{l|}{F2L}   & \multicolumn{1}{l|}{FL2FL} & \multicolumn{1}{l|}{F2FL} & FL2FL \\ \hline
\multicolumn{1}{|l|}{\multirow{2}{*}{\begin{tabular}[c]{@{}l@{}}Upper\\ Layer\end{tabular}}} & Epoch       & \multicolumn{1}{l|}{\textbf{39000}} & -     & \multicolumn{1}{l|}{26000} & -     & \multicolumn{8}{l|}{\multirow{2}{*}{}}                                                                                                                                                                        \\ \cline{2-6}
\multicolumn{1}{|l|}{}                                                                       & Time{[}s{]} & \multicolumn{1}{l|}{\textbf{1344}}  & -     & \multicolumn{1}{l|}{3616}  & -     & \multicolumn{8}{l|}{}                                                                                                                                                                                         \\ \hline
\multicolumn{1}{|l|}{\multirow{2}{*}{\begin{tabular}[c]{@{}l@{}}Lower\\ Layer\end{tabular}}} & Epoch       & \multicolumn{1}{l|}{\textbf{9000}}  & -     & \multicolumn{1}{l|}{3000}  & -     & \multicolumn{1}{l|}{-}    & \multicolumn{1}{l|}{-}     & \multicolumn{1}{l|}{-}    & \multicolumn{1}{l|}{-}     & \multicolumn{1}{l|}{49000} & \multicolumn{1}{l|}{-}     & \multicolumn{1}{l|}{-}    & -     \\ \cline{2-14} 
\multicolumn{1}{|l|}{}                                                                       & Time{[}s{]} & \multicolumn{1}{l|}{\textbf{3380}}  & -     & \multicolumn{1}{l|}{544}   & -     & \multicolumn{1}{l|}{-}    & \multicolumn{1}{l|}{-}     & \multicolumn{1}{l|}{-}    & \multicolumn{1}{l|}{-}     & \multicolumn{1}{l|}{75324} & \multicolumn{1}{l|}{-}     & \multicolumn{1}{l|}{-}    & -     \\ \hline
\end{tabular}
\end{table*}

\section{Discussion}
Although the ordinary LSTM could succeed in only the first experiment, the proposed method could execute subsequent tasks. These results demonstrate the effectiveness of the hierarchical structure.
In the third experiment, the trained lower layer succeeded in writing unlearned characters by receiving positional trajectories collected with direct teaching. However, performance improvement is expected when the diversity of the training datasets is increased in the data collection phase. In another method, data augmentation is considered.
In the experiments, the proposed method required less than 1/20 of the training time compared to conventional methods. 
In the comparative methods, more exponential training time is required for tasks that are more complex and require long-term inference. However, in the proposed method, the exponential increase in training time can be suppressed by using a structure with three or more levels of hierarchy.
Furthermore, we have succeeded in integrating the motion inference and images taken with a the camera, which works at a slow sampling rate. 
Hence, the proposed method can adapt to complex tasks that require multiple sensors with different sampling frequencies. However, detailed information is not described here because it is beyond the scope of this research. 
In addition, because the upper and lower layers are trained independently, they can be designed using different NN models. For example, a model with an upper layer provided by reinforcement learning and a lower layer characterized where by LSTM may be useful.
With this combination, the upper layer, which is trained with sim2real, generates various trajectories, and lower layer treats force information.
In the future, we will apply the proposed method to longer-term tasks by adding three or more levels of hierarchy to ensure that more complex tasks can be executed by reusing multiple trained lower layers.

\section{conclusion}
As an expansion of our past research, a bilateral control-based hierarchical imitation learning framework was proposed here. In our proposed method, the upper layer, whose timescale is large, and lower layer, whose timescale is short, are separately trained.
With this framework, robots can execute more long-term tasks while maintaining the advantages of bilateral control-based imitation learning, such as fast movement and force adjustment.

In experiment 2, the first proposed method required less than 1/20 of the training time compared to other hierarchical methods.
Additionally, the second proposed method demonstrated the best reproducibility.
In experiment 3, the lower layer of the proposed method, which was trained in experiment 2, succeeded in writing unlearned characters. This is because the lower layer learns the motion primitive and receives a rough positional trajectory.

In addition, in experiment 4, the proposed method required less than 1/20 of the training time compared to other hierarchical methods.
Moreover, although this task is the most difficult among all experiments because it requires 30 s, the proposed method exhibited the best success rate here.

As described above, the proposed method can be used for a difficult task that requires long-term inference, dynamic motion generation, and control of the contact force with the environment simultaneously.
Another feature of the proposed method is that it can be trained with a very short training time. Moreover, because it does not use inductive bias for a task, it is likely to be useful for general tasks.
In the future, we will apply our method to perform various tasks and verify its general usefulness.

\begin{comment}
{\BibTeX} does not work by magic. It doesn't get the bibliographic
data from thin air but from .bib files. If you use {\BibTeX} to produce a
bibliography you must send the .bib files. 
\end{comment}

%ダブルクオテーションのことこｒは、7で

\begin{IEEEbiography}[{\includegraphics[width=1in,height=1.25in,clip,keepaspectratio]{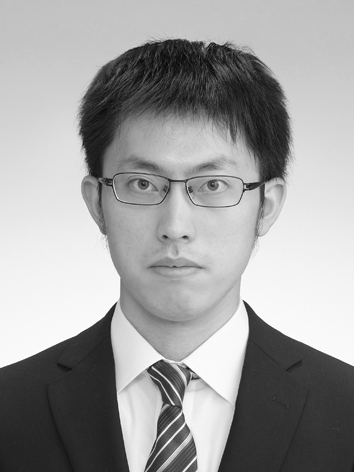}}]{Kazuki Hayashi}
received his B.E. degree in mechanical engineering from the Tokyo University of Agriculture and Technology, Tokyo, Japan, in 2020.
He is currently pursuing an M.E. degree from the Department of Information and Systems at the University of Tsukuba. His research interests include robotics, motion generation, and neural networks.
\end{IEEEbiography}

\begin{IEEEbiography}[{\includegraphics[width=1in,height=1.25in,clip,keepaspectratio]{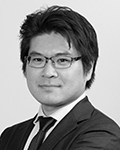}}]{Sho Sakaino} (Member, IEEE)
  received his B.E. degree in system design engineering and M.E. and Ph.D. degrees in integrated design engineering from Keio University, Yokohama, Japan, in 2006, 2008, and 2011, respectively. He was an assistant professor at Saitama University from 2011 to 2019. Since 2019, he has been an associate professor at the University of Tsukuba. His research interests include mechatronics, motion control, robotics, and haptics. He received the IEEJ Industry Application Society Distinguished Transaction Paper Award in 2011 and 2020. He also received the RSJ Advanced Robotics Excellent Paper Award in 2020.
\end{IEEEbiography}
\begin{IEEEbiography}[{\includegraphics[width=1in,height=1.25in,clip,keepaspectratio]{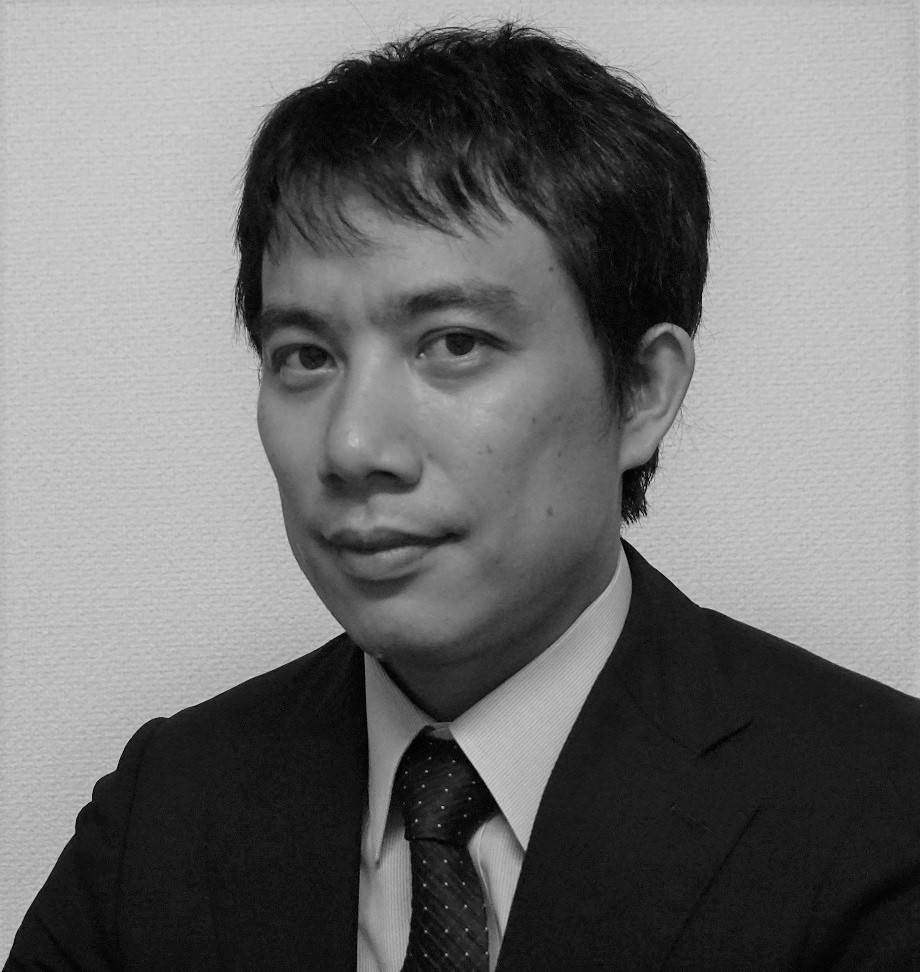}}]{Toshiaki Tsuji} (Senior Member, IEEE)
  received his B.E. degree in system design engineering and M.E. and Ph.D. degrees in integrated design engineering from Keio University, Yokohama, Japan, in 2001, 2003, and 2006, respectively. He was a Research Associate in the Department of Mechanical Engineering, Tokyo University of Science, from 2006 to 2007. He is currently an associate professor in the Department of Electrical and Electronic Systems, Saitama University, Saitama, Japan. His research interests include motion control, haptics, and rehabilitation robots. Dr. Tsuji received the FANUC FA and Robot Foundation Original Paper Award in 2007 and 2008, respectively. He also received the RSJ Advanced Robotics Excellent Paper Award and IEEJ Industry Application Society Distinguished Transaction Paper Award in 2020.
\end{IEEEbiography}

\EOD

\end{document}